%% file: 2025-uncertainty.tex
\pdfoutput=1

\documentclass[10pt, a4paper]{article}

\usepackage{lrec2026} 
\usepackage{amsmath}
\usepackage{soul}
\usepackage{comment}
\usepackage{caption}
\usepackage{subcaption}
\usepackage{booktabs} 
\usepackage{multirow} 
\usepackage{natbib} 
\usepackage{array}           
\usepackage{longtable}       
\usepackage{amsfonts} 
\usepackage{bbm}
\usepackage{covington}
\usepackage{todonotes}
\usepackage[all]{hypcap}
\usepackage{booktabs}
\usepackage{array}
\usepackage{tabularx}
\usepackage{threeparttablex}  
\usepackage{adjustbox}   
\usepackage{makecell}    
\newcommand{\fiteq}[1]{\adjustbox{max width=\linewidth}{$#1$}}

\usepackage{caption}
\usepackage{hyperref}
\usepackage[normalem]{ulem}
\newcommand{\tightpara}[1]

\usepackage{makecell}
\newcolumntype{L}[1]{>{\raggedright\arraybackslash}p{#1}}

\makeatletter
\renewcommand{\paragraph}{%
  \@startsection{paragraph}{4}%
  {\z@}{1.25ex \@plus 1ex \@minus .2ex}{-1em}%
  {\normalfont\normalsize\bfseries}%
}
\makeatother

\usepackage{makecell} 
\usepackage{amssymb}

\title{To Predict or Not to Predict? Towards reliable uncertainty estimation in the presence of noise}
\name{Nouran Khallaf, Serge Sharoff} 

\address{Centre for Translation, Localisation and Interpreting Studies \\
School of Languages, Cultures and Societies\\
University of Leeds, UK}

\abstract{
This study examines the role of uncertainty estimation (UE) methods in multilingual text classification under noisy and non-topical conditions. Using a complex-vs-simple sentence classification task across several languages, we evaluate a range of UE techniques against a range of metrics to assess their contribution to making more robust predictions. Results indicate that while methods relying on softmax outputs remain competitive in high-resource in-domain settings, their reliability declines in low-resource or domain-shift scenarios. In contrast, Monte Carlo dropout approaches demonstrate consistently strong performance across all languages, offering more robust calibration, stable decision thresholds, and greater discriminative power even under adverse conditions. We further demonstrate the positive impact of UE on non-topical classification: abstaining from predicting the 10\% most uncertain instances increases the macro F1 score from 0.81 to 0.85 in the Readme task. By integrating UE with trustworthiness metrics, this study provides actionable insights for developing more reliable NLP systems in real-world multilingual environments. See {\footnotesize \url{https://github.com/Nouran-Khallaf/To-Predict-or-Not-to-Predict}}
\\ \newline 
}

\begin{document}

\maketitleabstract

\section{Introduction}
\label{sec:introduction}
In many real-world NLP applications, it is not enough for a classifier to make predictions; users also need to know when the classifier is more likely to be wrong for an individual instance. This is particularly challenging in tasks involving non-topical classification, noisy data, and multilingual settings. In this paper, we focus on the task of sentence complexity classification, which often experiences performance degradation under cross-lingual transfer and domain shift.

To address these challenges, we investigate the role of Uncertainty Estimation (UE), i.e., the ability to quantify uncertainty of a model when making predictions on instances in a test set \citep{vazhentsev23uncertainty}. UE serves as a critical failure indicator by identifying cases where the model is uncertain and should abstain from making predictions, thereby improving overall robustness. While many UE approaches have recently been proposed, there has been no comprehensive evaluation across multiple languages, noisy contexts, and non-topical classification tasks.

This study makes the following contributions:
\vspace{-0.5ex}
\begin{enumerate}
\item \textbf{Comprehensive evaluation of UE methods:} We benchmark nine popular techniques, including probabilistic, distributional, geometric, and hybrid approaches.
\item \textbf{Multilingual and cross-domain investigation:} We evaluate these methods across seven languages (Arabic, Catalan, English, French, Hindi, Russian, and Spanish) and three datasets to assess their robustness to language and domain variation.
\item \textbf{Analysis of multiple evaluation metrics for UE assessment:} We compare nine widely used metrics, covering uncertainty discrimination, calibration, and selective prediction perspectives. We further analyse correlations among these measures within and across the perspectives. 
\item \textbf{Insights into selective prediction:} We show that abstaining from 5--10\% of the most uncertain predictions can yield significant improvements in macro F1 score.
\end{enumerate}

The remainder of the paper is organized as follows. Section~\ref{sec:methodology} describes the methodology, including the dataset, and  the experimental setup. Section~\ref{sec:uncertainty_measures} describes the UE methods, and the evaluation metrics. Section~\ref{sec:results} presents the results across the multilingual datasets.

\section{Methodology}
\label{sec:methodology}
\subsection{Datasets \label{secData}}

Table~\ref{tab:corpus} lists the datasets used in this study.  As our focus is on multilingual detection of segments which are difficult to read, our training dataset is Readme++, a multilingual collection of paragraphs graded by their CEFR level \cite{naous24readme}.  We converted it to a binary task: Simple is B1 or below, Complex is C1 or C2. 

In addition to testing the multilingual model on its original dataset via cross-validation, we also test it on other datasets to assess its robustness across languages and domains. One test set comes from Vikidia,\footnote{\url{https://www.vikidia.org/}} a website that maintains Wikipedia-style content aimed at ``children and anyone seeking easy-to-read content''. We have removed the Vikidia entries marked as stubs (with little content at the moment) and collected the corresponding main Wikipedia entries for the respective languages.  This removes topic biases, which often lead to unreasonable accuracy scores in non-topical classification tasks.  The set of languages for testing on the Vikidia corpus is determined by the availability of annotators for quality control purposes.  As another test set we use Simplext, a manually simplified set of Spanish news articles from the domains of national and international news, society, and culture \cite{simplext2015}.

\begin{table}[t]
\centering
\small
\caption{Statistics of the datasets. }
\label{tab:corpus}
\begin{tabular}{l|rr|rr}
\toprule
 & \multicolumn{2}{c|}{\textbf{Simple}} & \multicolumn{2}{c}{\textbf{Complex}} \\
\textbf{Language} & \textbf{\#Ex's} & \textbf{IQR} & \textbf{\#Ex's} & \textbf{IQR} \\
\midrule
\multicolumn{5}{l}{\textbf{ReadMe}} \\
Arabic  & 766  & 9--20  & 533  & 12--33 \\
English & 1,296 & 9--20  & 448  & 13--32 \\
French  & 810  & 10--23 & 367  & 14--35 \\
Hindi   & 569  & 9--20  & 396  & 11--29 \\
Russian & 702  & 9--20  & 328  & 10--26 \\

\midrule
\multicolumn{5}{l}{\textbf{Vikidia / Wikipedia}} \\
Catalan & 1,236 & 12--27 & 1,397 & 18--38 \\
English & 30,553 & 11--23 & 26,790 & 15--32 \\
French  & 41,127 & 13--29 & 41,127 & 16--36 \\
Italian & 32,994 & 12--26 & 34,283 & 13--32 \\
Spanish & 4,038  & 13--29 & 4,038  & 18--38 \\
\midrule
\multicolumn{5}{l}{\textbf{Simplext}} \\
Spanish & 167 & 11--17 & 167 & 16--42 \\
\bottomrule
\end{tabular}%
\end{table}

Table \ref{tab:corpus} shows the number of examples for each language and dataset in each category, as well as the interquartile range (IQR) of segment lengths, as this is another potential confounding factor in non-topical text classification.

\subsection{Classifiers and Experimental Setup}

We trained a multilingual classification model using multilingual Bert \citep{devlin2019bert} with 5-fold cross-validation to estimate the robustness of the classifiers and the respective UE scores. Preliminary experiments with XLM-Roberta showed very similar trends; therefore, we report only mBERT below.  Recent studies demonstrate that larger generative LLMs such as GPT or LLaMA do not consistently outperform BERT-sized PLMs in text classification tasks \cite{edwards24incontext}, as was also shown in our own experiments on text complexity \cite{khallaf25mtsummit}.  Therefore, we focus on more computationally efficient PLMs in our task.

In terms of hyper-parameters, we fine-tune using the AdamW optimizer with a learning rate of $5 \times 10^{-5}$, a cosine decay scheduler (with 10\% linear warmup), and early stopping (patience = 5). To promote generalisation and training stability, we apply dropout ($p = 0.3$), mixed-precision training (FP16), and gradient clipping to prevent excessively large gradient updates during unstable training steps, which can occur more frequently with imbalanced or low-resource data.

\section{Uncertainty Evaluation}
\label{sec:uncertainty_measures}

\begin{table*}[t]
\centering
\small
\caption{Summary of Uncertainty Estimation methods, grouped by the underlying approaches.}
\label{tab:uncertainty_formulations_grouped}
\resizebox{\textwidth}{!}{%
\begin{tabular}{l l l}
\toprule
\textbf{Method} & \textbf{Formulation} & \textbf{Key Characteristics} \\
\midrule

\multicolumn{3}{l}{\textbf{Class probability-based methods}} \\
\midrule
\textbf{SR} &
Top-class softmax probability. &
Directly available, but calibration is questionable.
\\

\textbf{SMP} &
MC Dropout-averaged top-class probability. &
Captures disagreement across stochastic passes. \\


\textbf{ENT} &
Entropy over softmax probabilities. &
Captures prediction ambiguity. \\

\textbf{ENT-MC} &
MC Dropout-averaged entropy over probabilities. &
Captures ambiguity across stochastic passes. \\

\textbf{PV} &
MC Dropout-averaged variance in probabilities. &
Captures disagreement across stochastic passes. \\

\textbf{BALD} &
Mutual information between model and predictions. &
Isolates epistemic uncertainty. \\

\midrule
\multicolumn{3}{l}{\textbf{Feature-Based Methods}} \\
\midrule

\textbf{MD} &
Distance of test embeddings. &
Captures distance from the training distribution. \\

\textbf{ISOF} &
Isolation-based anomaly detection on embeddings. &
Finds OOD inputs using tree partitions. \\

\textbf{LOF} &
Local density deviation on embeddings. &
Finds OOD inputs using local density. \\

\midrule
\multicolumn{3}{l}{\textbf{Hybrid Methods}} \\
\midrule

\textbf{HUQ-MD} &
Rank-based mix of MD and Mean probability. &
Balances model and data uncertainty. \\

\bottomrule
\end{tabular}
}
\end{table*}

\subsection{Uncertainty Estimation Methods}

We evaluate UE methods belonging to three groups of approaches: probabilistic, feature geometry, as well as rank-based hybrids of geometric and probabilistic approaches, see Table~\ref{tab:uncertainty_formulations_grouped}. Formal definitions are listed in the Appendix.

\paragraph{Softmax Response (SR)} is a probability-based method that quantifies uncertainty using the maximum predicted class probability from the softmax layer, which is already computed during each inference \citep{Guo2017}.

\paragraph{Sampled Max Probability (SMP)} is a probability-based method that extends SR by introducing randomness at inference time using MC Dropout  \citep{gal2016dropout}. Instead of relying on a single softmax output, SMP averages the class probabilities over \( T \) forward passes with dropout, then computes uncertainty using the maximum of the mean probabilities \citep{shelmanov-etal-2021-certain}.

\paragraph{Entropy (ENT)} quantifies uncertainty by measuring the spread of the predicted probability distribution 
\citep{ovadia2019can}. ENT can be computed directly from the softmax output of a single forward pass, or using averaged probabilities over multiple stochastic passes (ENT-MC). A higher entropy value indicates a prediction which is more uniform over the predicted classes (and thus more uncertain).

\paragraph{Probability Variance (PV)} quantifies epistemic uncertainty by checking how much the predicted class probabilities change across \(T\) MC–Dropout runs. Unlike ENT-MC or SMP, which rely on the average prediction, PV looks at the spread of those predictions to reflect model instability: bigger spread means the model is more uncertain \citep{lakshminarayanan2017}. 

\paragraph{Bayesian Active Learning by Disagreement (BALD)} quantifies epistemic uncertainty using information gain in terms of predictive entropies \citep{houlsby2011bald}.  Variance is approximated via MC-dropout \citep{gal2016dropout}.

\paragraph{Mahalanobis Distance (MD)} estimates uncertainty from the model’s representations rather than output probabilities. It compares the embeddings of a test sample against the training data. Larger distances mean the test instance looks unlike what has been seen during training \citep{lee2018simple}.

\paragraph{Hybrid Uncertainty Quantification (HUQ–MD)} combines epistemic and aleatoric uncertainty \citep{vazhentsev23uncertainty}. 
Epistemic uncertainty comes from Mahalanobis Distance (MD). 
Aleatoric uncertainty comes from the model’s predicted probability for the chosen class: lower confidence (closer to a tie) means higher aleatoric uncertainty. 
HUQ ranks each test instance by both signals within the dataset and then combines the two ranks. 
This penalizes both overconfident mistakes and genuinely ambiguous cases.

\paragraph{Local Outlier Factor (LOF)} detects uncertainty by measuring how isolated a test example is within its local neighbourhood of training data: outliers are in a less dense region \citep{breunig2000lof}. 

\paragraph{Isolation Forest (ISOF)}  identifies examples that look unusual compared to the training data by building a set of binary trees that split the feature space; needing fewer splits to build a tree implies a more anomalous example \citep{liu2008isolation}. 

\subsection{Uncertainty evaluation metrics}

We organise our evaluation of uncertainty estimators in three groups of metrics (see Appendix~\ref{app:metrics_eval} for the formal definitions):

\begin{itemize}
\item \textbf{Uncertainty discrimination} Ranking ability of uncertainty estimates to distinguish correct from incorrect predictions \citep{ovadia2019can}. High discrimination means the model is better at separating correct vs. incorrect predictions by assigning higher uncertainty to errors.
\item \textbf{Calibration Metrics} evaluate whether the model’s predicted confidence (the inverse of uncertainty) reflects the true likelihood of correctness \citep{Guo2017, naeini15calibration}.
\item \textbf{Selective Prediction Metrics} measure how effectively uncertainty can improve reliability by rejecting uncertain predictions \citep{geifman2017selective}. 
\end{itemize}

When a quality metric requires \emph{confidence} rather than uncertainty, we derive it as the opposite to normalised uncertainty ($c_i = 1 - u_i$).  

\begin{table*}[t]
\centering
\small 
\caption{Summary of UE quality metrics grouped by the evaluation perspective.}
\label{tab:evaluation_metrics_grouped}
\setlength{\tabcolsep}{2pt}
\begin{tabular}{l l p{0.85\textwidth}}
\toprule
\textbf{Metric} & \textbf{Range} & \textbf{Interpretation} \\
\midrule

\multicolumn{3}{l}{\textbf{Uncertainty Discrimination}} \\
\midrule
\textbf{ROC-AUC} & [0, 1] & Indicates whether incorrect predictions have higher uncertainty. Higher is better. \\
\textbf{AU-PRC} & [0, 1] & Identifies incorrect predictions via uncertainty. Higher is better. \\

\midrule
\multicolumn{3}{l}{\textbf{Calibration Metrics}} \\
\midrule
\textbf{C-Slope} & [0, $\infty$] & Regression of accuracy on confidence. $<$1 = underconfident; $>$1 = overconfident. \\
\textbf{CITL} & $[-1, 1]$ & Measures the difference between confidence and accuracy. The best value is 0. \\
\textbf{ECE} & [0, 1] & Expected calibration error. Lower is better. \\

\midrule
\multicolumn{3}{l}{\textbf{Selective Prediction Metrics}} \\
\midrule
\textbf{RC-AUC} & [0, 1] & Area under the risk-coverage curve. Lower is better. \\
\textbf{N.RC-AUC} & [0, 1] & Normalised Area under the risk-coverage curve. Higher is better. \\
\textbf{E-AUoptRC} & [0, 1] & Area under the RC curve up to the full-set performance threshold. Lower is better. \\

\textbf{Trust Index} & [0, 1] & Performance of macro-F1 on the most confident $c^*$ fraction of predictions. Higher is better. \\

\bottomrule
\end{tabular}
\end{table*}
\begin{table*}[!t]
\centering
\setlength{\tabcolsep}{2pt}
\small
\caption{Macro F1 with standard deviations across the folds. The source dataset for training the classifier is Readme.  The domain shift tested on Vikidia/Wikipedia and Simplext (Spanish only)}
\label{tab:full_metrics_with_accuracy}
\resizebox{\textwidth}{!}{%
\begin{tabular}{c|c}

\begin{tabular}{llrrr} \multicolumn{5}{l}{\textbf{Readme}}\\ Language & Class & P \hspace{0.2cm} $\pm$ Std & R \hspace{0.2cm} $\pm$ Std & F1 \hspace{0.2cm} $\pm$ Std\\ \midrule \textbf{Arabic} & complex & 0.77 $\pm$ 0.050 & 0.86 $\pm$ 0.058 & 0.81 $\pm$ 0.035\\ & simple & 0.90 $\pm$ 0.023 & 0.81 $\pm$ 0.067 & 0.85 $\pm$ 0.034\\ \textbf{English} & complex & 0.80 $\pm$ 0.056 & 0.79 $\pm$ 0.080 & 0.79 $\pm$ 0.036\\ & simple & 0.93 $\pm$ 0.022 & 0.93 $\pm$ 0.029 & 0.93 $\pm$ 0.013\\ \textbf{French} & complex & 0.79 $\pm$ 0.047 & 0.85 $\pm$ 0.084 & 0.81 $\pm$ 0.037\\ & simple & 0.93 $\pm$ 0.032 & 0.89 $\pm$ 0.047 & 0.91 $\pm$ 0.021\\ \textbf{Hindi} & complex & 0.79 $\pm$ 0.048 & 0.74 $\pm$ 0.128 & 0.76 $\pm$ 0.075\\ & simple & 0.83 $\pm$ 0.069 & 0.86 $\pm$ 0.044 & 0.84 $\pm$ 0.033\\ \textbf{Russian} & complex & 0.75 $\pm$ 0.083 & 0.78 $\pm$ 0.113 & 0.75 $\pm$ 0.057\\ & simple & 0.90 $\pm$ 0.037 & 0.87 $\pm$ 0.067 & 0.88 $\pm$ 0.028\\ \midrule \multicolumn{2}{l}{\textbf{Macro averaged}} & 0.84 $\pm$ 0.013 & 0.84 $\pm$ 0.016 & \textbf{0.83} $\pm$ 0.014\\

\bottomrule
\end{tabular}
     &  
\begin{tabular}{llrrr}
\multicolumn{5}{l}{\textbf{Vikidia/Wikipedia}}\\
Language & Class & P \hspace{0.2cm} $\pm$ Std & R \hspace{0.2cm} $\pm$ Std & F1 \hspace{0.2cm} $\pm$ Std\\
\midrule

\textbf{Catalan} & complex & 0.72 $\pm$ 0.044 & 0.49 $\pm$ 0.137 & 0.57 $\pm$ 0.101\\
 & simple & 0.62 $\pm$ 0.044 & 0.80 $\pm$ 0.075 & 0.69 $\pm$ 0.017\\
\textbf{English} & complex & 0.79 $\pm$ 0.020 & 0.62 $\pm$ 0.026 & 0.69 $\pm$ 0.021\\
 & simple & 0.68 $\pm$ 0.016 & 0.83 $\pm$ 0.017 & 0.75 $\pm$ 0.015\\
\textbf{French} & complex & 0.69 $\pm$ 0.006 & 0.60 $\pm$ 0.008 & 0.64 $\pm$ 0.006\\
 & simple & 0.64 $\pm$ 0.005 & 0.73 $\pm$ 0.008 & 0.68 $\pm$ 0.005\\
\textbf{Italian} & complex & 0.66 $\pm$ 0.017 & 0.64 $\pm$ 0.102 & 0.65 $\pm$ 0.058\\
 & simple & 0.66 $\pm$ 0.049 & 0.68 $\pm$ 0.048 & 0.66 $\pm$ 0.012\\
\textbf{Spanish} & complex & 0.68 $\pm$ 0.017 & 0.72 $\pm$ 0.007 & 0.70 $\pm$ 0.011\\
 & simple & 0.71 $\pm$ 0.011 & 0.66 $\pm$ 0.024 & 0.68 $\pm$ 0.017\\
\midrule
\multicolumn{2}{l}{\textbf{Macro averaged}} & 0.69 $\pm$ 0.035 & 0.68 $\pm$ 0.036 & \textbf{0.68} $\pm$ 0.040\\
\bottomrule
\end{tabular}

\end{tabular}
}
\resizebox{\columnwidth}{!}{%
\begin{tabular}{llrrr}

\multicolumn{5}{l}{\textbf{Simplext–Spanish}}\\
Language & Class & P \hspace{0.2cm} $\pm$ Std & R \hspace{0.2cm} $\pm$ Std & F1 \hspace{0.2cm} $\pm$ Std\\
\midrule
\textbf{Spanish} & complex & 0.79 $\pm$ 0.152 & 0.54 $\pm$ 0.174 & 0.62 $\pm$ 0.112\\ \textbf{Simplext} & simple & 0.65 $\pm$ 0.064 & 0.83 $\pm$ 0.150 & 0.72 $\pm$ 0.067\\ \midrule \multicolumn{2}{l}{\textbf{Macro averaged}} & 0.72 $\pm$ 0.108 & 0.69 $\pm$ 0.162 & \textbf{0.67} $\pm$ 0.090 \\

\bottomrule
\end{tabular}
}
\end{table*}
\subsubsection{Uncertainty discrimination}
\paragraph{Receiver Operating Characteristic- Area Under the Curve (ROC-AUC)} measures the discriminative ability of an uncertainty score, i.e. whether the incorrect predictions have higher uncertainty. We treat prediction correctness as the positive class ($y_i=1$ if the prediction is correct, $0$ otherwise). Higher ROC-AUC indicates that correct predictions receive higher confidence.

\paragraph{Area Under Precision-Recall Curve (AU-PRC)} is similar to ROC-AUC, but it uses the Precision-Recall curve, it also treats incorrect predictions as the positive class. This metric is claimed to be particularly informative for higher-accuracy models \citep{davis2006relationship}.

\subsubsection{Calibration metrics}
\paragraph{Calibration Slope (C-slope)} evaluates how well the confidence score aligns with prediction accuracy. It is computed by fitting a linear regression model between the \textbf{confidence} and the correctness indicator.

\paragraph{Calibration-in-the-large (CITL)} captures the average bias in the confidence score. \emph{CITL} $=0$ is ideal; positive values indicate the model is {overconfident}, and negative values indicate {underconfident} \citep{naeini15calibration, kumar19calibration}.

\paragraph{Expected Calibration Error (ECE)} measures how well predicted confidences match observed accuracy \citep{ao23failure}. 
Predictions are grouped into confidence bins; within each bin, we compute the absolute gap between mean confidence and empirical accuracy, then take a (sample-weighted) average over bins. 
Lower ECE indicates better calibration; it is important to note that the values depend on the chosen binning strategy. We compute ECE with $15$ equal-width confidence bins over $[0,1]$ and take a sample-weighted average of the absolute gap between mean confidence and empirical accuracy within each bin.

\subsubsection{Selective prediction}
\paragraph{Risk-Coverage Area Under Curve (RC-AUC):}
RC-AUC assesses how prediction risk accumulates as we increase the fraction of retained (non-rejected) predictions.
We rank samples by decreasing confidence $c_i$ and define risk using the error indicator $r_i=\mathbbm{1}[\hat{y}_i\neq y_i]$.
Lower RC-AUC indicates that the most confident predictions are less risky.

\paragraph{Normalised RC-AUC:}
\citet{vazhentsev2025uncertainty} claim that the RC-AUC absolute values are not normalised, so they are dataset- and model-dependent, which makes them difficult to interpret in isolation across the models and the datasets.  They suggest rescaling RC-AUC between a random ranking (worst reasonable baseline) and an oracle ranking by true risk (best case). 


\begin{figure*}[ht]
  \centering
  \includegraphics[width=1\linewidth]{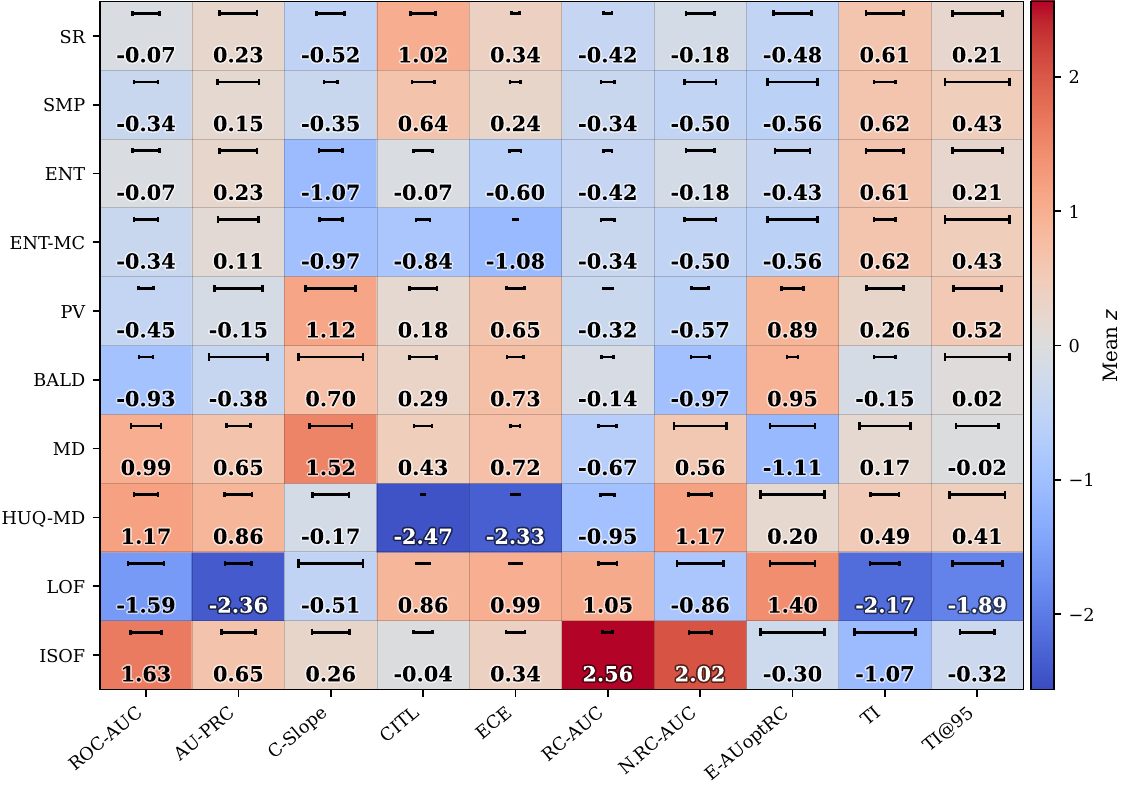}
  \caption{Cross-language average $z$-scores after applying a direction-aware benefit transform across languages.  
  Cell colors indicate relative performance (better $\rightarrow$ warmer), and horizontal whiskers show the standard deviation of $z$ across languages.  }
  \label{fig:agg_meanz_whiskers}
\end{figure*}
\paragraph{Expected Area-Under-Optimal-Risk-Coverage} (E-AUoptRC) measures the area under the risk-coverage curve up to the accuracy of the model, corresponding to the optimal coverage point beyond which further rejections do not improve performance \citep{geifman2017selective}. In our setting, $c^*$ is defined by the model's macro-F$_1$ (rather than accuracy).

\paragraph{Trust Index (TI).}
TI evaluates performance when keeping only the most confident predictions \citep{ao23failure}. 
We adapt the original formulation with accuracy to macro-F$_1$ to handle class imbalance. 
Let \(n\) be the number of test samples and \(c^*\) the coverage at which the model’s macro-F$_1$ on the full set is reached. 
This involves ranking the predictions by confidence (descending), keeping the top \(k=\lfloor c^*\!\cdot n\rfloor\), and measuring macro-F$_1$ on this subset. 
We report two settings: the optimal selective threshold (often \(c^*\!\approx 0.50\)) and \(c=0.95\) (abstaining from prediction on the least-confident 5\%).

\section{Results and Error Analysis}
\label{sec:results}
\subsection{Evaluation of the classifier}
Table \ref{tab:full_metrics_with_accuracy} reports the quality of the baseline classifier for each language via cross-validation with the respective standard deviation values across the folds. Overall, the macro F1 score is acceptable.  However, the classifiers have a greater problem in detection of more complex examples, this is true across all languages, irrespectively of the amount of data for pre-training the mBert model. In terms of the domain and language shift, there is a consistent drop, in both cases when the classifier is applied to a new domain with the languages present in the training set (English and French) and when it is applied to a new language (Catalan, Italian and Spanish). Now we want to understand how different UE scores impact classification and how this is influenced by the domain and language shift.

\subsection{Uncertainty Estimation scores}
The heatmap in Figure~\ref{fig:agg_meanz_whiskers} aggregates uncertainty-evaluation results across languages by converting each metric to a benefit score on a shared ``higher-is-better'' axis before computing $z$-scores. 
For metrics where larger values are better (e.g., ROC-AUC, AU-PRC, TI), we use the raw values; for metrics where smaller values are better (e.g., ECE, RC-AUC, E-AUoptRC), we invert the sign so that higher always indicates better performance; and for metrics with an ideal target value (e.g., CITL with target $0$), we score systems by closeness to that target (e.g., $-\lvert \mathrm{CITL}\rvert$). The transformed scores are then standardised into $z$-scores to enable direct comparison across methods and languages. Colour intensity reflects relative performance; horizontal whiskers indicate variability across folds and languages. Per-language values appear in Appendix in the full version of the paper.

Our first observation is the inconsistency of UE scores across the quality metrics. One of the surprises is that while the softmax classifiers have been claimed to be poorly calibrated, as they seem to output high probabilities for incorrect predictions \citep{Guo2017}, our experiment shows that SR is a strong measure according to nearly all of the quality metrics across the tested languages,  achieving the best CITL scores with relatively low variance across the folds and languages (shorter whiskers).  SMP (an MC-dropout version of SR) only shows clear advantage over SR in TI@95 scores.

Dropout-based approaches (PV, BALD) show strong calibration according to C-Slope and ECE, particularly in English, where C-slope reaches 0.75 and 0.72 and ECE $\approx 0.08$, but again with higher variance across the folds. 

Older outlier detection methods (ISOF, LOF) often achieve the highest discrimination and selective prediction scores in many settings, in particular the best scores for ROC-AUC, RC-AUC and N.RC-AUC.  For instance, in the per-language results (Appendix Table~\ref{tab:metrics_summary}), \textit{ISOF} attains ROC-AUC $\approx 0.73$ (AR), $0.74$ (RU), and $0.67$ (HI), yet its aggregate stability is weaker. This instability highlights their dataset-dependent behaviour. 

Overall, \textit{MD} is the most reliable and consistent choice across languages and evaluation lenses. The hybrid score \textit{HUQ\mbox{-}MD} which combines MD and SR uncertainty is consistently good on discrimination and selection quality measures, but its calibration is much weaker. \textit{ISOF} is often the best or near-best for discrimination/selection but is unstable across languages. Finally, the simple \textit{SR} and \textit{ENT} baselines remain surprisingly effective and robust across languages, offering competitive performance at no additional computational cost.

\begin{table}[!t]
\setlength{\tabcolsep}{3pt}
\centering
\small
\caption{Kendall's $\tau$ within metric perspectives.}
\label{tab:metric_correlations_kendall}
\resizebox{\columnwidth}{!}{%
\begin{tabular}{l r r r r r}
\toprule
\textbf{Metric Pair} & \textbf{AR} & \textbf{EN} & \textbf{FR} & \textbf{HI} & \textbf{RU} \\
\midrule
\multicolumn{6}{l}{\textbf{Uncertainty Discrimination}} \\
AU-PRC vs ROC-AUC & \textbf{0.52} & \textbf{0.61} & \textbf{0.65} & \textbf{0.53} & \textbf{0.58} \\
\midrule
\multicolumn{6}{l}{\textbf{Calibration Metrics}} \\
CITL vs ECE & \textbf{-0.39} & \textbf{-0.61} & \textbf{-0.55} & \textbf{-0.27} & \textbf{-0.32} \\
C-Slope vs CITL & \textbf{0.31} & \textbf{0.42} & \textbf{0.33} & -0.03 & 0.11 \\
C-Slope vs ECE & \textbf{-0.37} & \textbf{-0.46} & \textbf{-0.45} & \textbf{-0.28} & \textbf{-0.29} \\
\midrule
\multicolumn{6}{l}{\textbf{Selective Prediction Metrics}} \\
E-AUoptRC vs RC-AUC & -0.08 & \textbf{-0.35} & \textbf{-0.34} & \textbf{-0.40} & \textbf{-0.30} \\
E-AUoptRC vs NRC-AUC & \textbf{-0.33} & \underline{-0.22} & \textbf{-0.46} & \textbf{-0.44} & \textbf{-0.54} \\
E-AUoptRC vs TI & -0.07 & \underline{-0.21} & -0.16 & \textbf{-0.38} & \underline{-0.21} \\
NRC-AUC vs TI & \textbf{0.39} & 0.03 & 0.13 & \textbf{0.40} & 0.04 \\
RC-AUC vs TI & \textbf{0.43} & \underline{0.23} & \textbf{0.31} & \textbf{0.55} & \textbf{0.34} \\
RC-AUC vs NRC-AUC & \textbf{0.41} & \textbf{0.47} & \textbf{0.38} & \textbf{0.45} & \textbf{0.29} \\
E-AUoptRC vs TI@95 & \underline{-0.24} & \underline{-0.20} & -0.16 & \textbf{-0.31} & -0.19 \\
NRC-AUC vs TI@95 & \textbf{0.33} & -0.09 & 0.08 & \textbf{0.32} & 0.02 \\
RC-AUC vs TI@95 & \textbf{0.42} & 0.11 & \textbf{0.27} & \textbf{0.47} & \textbf{0.28} \\
TI vs TI@95 & \textbf{0.60} & \textbf{0.64} & \textbf{0.67} & \textbf{0.76} & \textbf{0.64} \\

\bottomrule
\end{tabular}
}
\end{table}

\begin{table}[!t]
\setlength{\tabcolsep}{3pt}
\centering
\small
\caption{Kendall's $\tau$ across metric perspectives.}
\label{tab:all_metric_correlations_kendall}
\resizebox{\columnwidth}{!}{%
\begin{tabular}{l r r r r r}
\toprule
\textbf{Metric Pair} & \textbf{AR} & \textbf{EN} & \textbf{FR} & \textbf{HI} & \textbf{RU} \\
\midrule
\multicolumn{6}{l}{\textbf{Discrimination vs Calibration}} \\
AU-PRC vs C-Slope & \textbf{0.45} & \textbf{0.47} & \textbf{0.31} & \textbf{0.51} & \textbf{0.57} \\
AU-PRC vs CITL & -0.02 & 0.05 & -0.16 & \underline{-0.22} & -0.09 \\
AU-PRC vs ECE & -0.03 & -0.07 & 0.03 & 0.00 & -0.03 \\
C-Slope vs ROC-AUC & \textbf{0.38} & \textbf{0.40} & \textbf{0.31} & \textbf{0.59} & \textbf{0.46} \\
CITL vs ROC-AUC & -0.14 & -0.01 & -0.14 & \underline{-0.24} & -0.13 \\
ECE vs ROC-AUC & -0.15 & -0.02 & 0.00 & -0.17 & -0.08 \\
\midrule
\multicolumn{6}{l}{\textbf{Discrimination vs Selection}} \\
AU-PRC vs E-AUoptRC & -0.16 & \underline{-0.23} & \textbf{-0.32} & \underline{-0.23} & -0.10 \\
AU-PRC vs NRC-AUC & \textbf{0.30} & \textbf{0.50} & \textbf{0.47} & \textbf{0.41} & \underline{0.22} \\
AU-PRC vs RC-AUC & 0.18 & \textbf{0.31} & \textbf{0.40} & 0.12 & 0.15 \\
AU-PRC vs TI & \textbf{0.34} & 0.15 & \textbf{0.53} & 0.16 & 0.14 \\
AU-PRC vs TI@95 & \underline{0.21} & 0.01 & \textbf{0.36} & -0.00 & -0.08 \\
E-AUoptRC vs ROC-AUC & \textbf{-0.37} & \underline{-0.23} & \textbf{-0.45} & \textbf{-0.42} & \textbf{-0.39} \\
NRC-AUC vs ROC-AUC & \textbf{0.72} & \textbf{0.75} & \textbf{0.76} & \textbf{0.80} & \textbf{0.57} \\
RC-AUC vs ROC-AUC & \textbf{0.34} & \textbf{0.55} & \textbf{0.41} & \textbf{0.37} & \textbf{0.32} \\
ROC-AUC vs TI & \textbf{0.44} & 0.16 & \textbf{0.30} & \textbf{0.41} & \underline{0.23} \\
ROC-AUC vs TI@95 & \textbf{0.34} & 0.01 & 0.18 & \textbf{0.28} & 0.07 \\
\midrule
\multicolumn{6}{l}{\textbf{Calibration vs Selection}} \\
C-Slope vs E-AUoptRC & \textbf{-0.32} & \textbf{-0.33} & \underline{-0.24} & \textbf{-0.37} & -0.17 \\
C-Slope vs NRC-AUC & \textbf{0.26} & \textbf{0.30} & 0.19 & \textbf{0.50} & 0.16 \\
C-Slope vs RC-AUC & \underline{0.19} & \textbf{0.35} & \underline{0.20} & \underline{0.25} & 0.12 \\
C-Slope vs TI & \textbf{0.30} & \underline{0.20} & \underline{0.25} & \textbf{0.31} & \underline{0.23} \\
C-Slope vs TI@95 & \textbf{0.30} & 0.10 & 0.18 & 0.17 & 0.02 \\
CITL vs E-AUoptRC & -0.10 & -0.13 & 0.05 & -0.01 & 0.00 \\
CITL vs NRC-AUC & \underline{-0.19} & -0.05 & -0.17 & \textbf{-0.29} & -0.16 \\
CITL vs RC-AUC & -0.06 & 0.06 & -0.14 & -0.12 & -0.03 \\
CITL vs TI & -0.13 & -0.06 & -0.19 & -0.12 & -0.02 \\
CITL vs TI@95 & -0.02 & -0.12 & \underline{-0.19} & -0.12 & -0.13 \\
E-AUoptRC vs ECE & \textbf{0.39} & 0.18 & 0.10 & \textbf{0.32} & \underline{0.23} \\
ECE vs NRC-AUC & \underline{-0.19} & 0.01 & 0.03 & -0.16 & -0.12 \\
ECE vs RC-AUC & -0.07 & 0.01 & 0.06 & -0.11 & 0.08 \\
ECE vs TI & -0.11 & -0.03 & -0.01 & -0.15 & -0.10 \\
ECE vs TI@95 & \underline{-0.24} & -0.04 & -0.04 & -0.09 & -0.07 \\

\bottomrule
\end{tabular}
}
\end{table}

\subsection{Efficiency estimates}
We report inference-time overhead per uncertainty method. Experiments were executed on an NVIDIA L40S  GPU with  48\, GB VRAM. In our implementation, the fastest methods are the feature-based OOD detectors LOF  ($\approx$1.34s per fold) and ISOF ($\approx$1.35s). Class probability-based scores (SR and ENT) are slightly slower ($\approx$1.60s per fold) and show higher variability across the folds (much longer for longer examples).  However, they come for free when the prediction task is needed anyway.  Mahalanobis-based scoring (MD and HUQ-MD) is moderately more expensive ($\approx$5.78s per fold), with HUQ-MD closely matching MD because it adds only a lightweight rank-combination step. Finally, MC-dropout probability-based methods (SMP, PV, BALD, ENT\_MC; $T{=}20$) have the highest runtime overhead ($\approx$21.12s per fold), dominated by repeated stochastic forward passes; their near-identical runtimes reflect that they share the same MC-dropout prediction block and differ mainly in inexpensive post-processing.

\subsection{Comparison of quality metrics}
Tables~\ref{tab:metric_correlations_kendall} and \ref{tab:all_metric_correlations_kendall} compare how the different quality evaluation metrics relate to each other by computing Kendall's $\tau$ correlations on the concatenated folds for each language.  Bold if $p<0.01$, underline if $p<0.05$. The use of  Kendall's $\tau$ rank-based correlation is justified by the need to compare non-linear patterns in the quality metrics.

Within each perspective (Table~\ref{tab:metric_correlations_kendall}), the discrimination (ROC-AUC and AU-PRC) and  calibration (C-Slope, CITL and ECE) metrics measure very similar properties of the UE scores within each other.  Correlation of AU-PRC with ROC-AUC shows that detection of errors (AU-PRC) correlates with ranking of positives and negatives errors (ROC-AUC). ROC-AUC is slightly more stable across the folds and languages (as shown in Figure \ref{fig:agg_meanz_whiskers}).

The similarity of calibration measures differs considerably across the languages.  The datasets for Hindi and Russian have been developed for the same purpose in the same team.  Also the performance of the respective classifiers is consistent for these languages (see Table~\ref{tab:full_metrics_with_accuracy}).  Nevertheless, CITL differs greatly in their case, which indicates limited robustness of this UE quality metric.  ECE is the most stable measure in this group.

The selective prediction evaluation measures show the greatest variation. Only closely related measures (RC-AUC vs N.RC-AUC and TI vs TI@95) show significant correlation across all languages, while there is little agreement in others, for example, N.RC-AUC correlates well with TI in AR/HI but it is close to zero in EN and RU.

Across the perspectives (Table~\ref{tab:all_metric_correlations_kendall}), there is little agreement with some exceptions.  For the Discrimination vs Selection perspectives, N.RC-AUC and ROC-AUC correlate strongly, suggesting that normalised rejection coverage is quite close to ROC-AUC in identifying well-separated predictions. ROC-AUC correlates with other Selection measures such as E-AUoptRC slightly better than AU-PRC, which is a reason to adopt it as the primary evaluation measure. 

Calibration tends to disagree with either discrimination or selection if all languages are considered, except for C-Slope, which does demonstrate statistically significant correlations with ROC-AUC and AU-PRC (Discrimination) and TI (Selection).

The Selective Prediction metrics are inherently different between each other and across the languages.  

\subsection{Selection at low rejection rates}
\label{sec:ue_low_rejection}
While Selective Prediction metrics vary, they are inherently interesting from the practical viewpoint because of their direct impact on robustness.  Here we want to investigate two questions:
\begin{enumerate}
\item the ability of the UE scores to improve prediction quality by abstaining from the decision;
\item the ability of metrics to capture this improved prediction quality.
\end{enumerate}

Metrics for evaluating selective predictions often place emphasis on the best rejection range.  In practice, however, it is more realistic to consider low rejection thresholds, since abstaining from predicting a large fraction of the test set is impractical.  

\begin{figure*}[!t]
  \centering
  \begin{subfigure}{\textwidth}
    \centering
    \includegraphics[width=\linewidth,trim=2pt 2pt 2pt 2pt,clip]{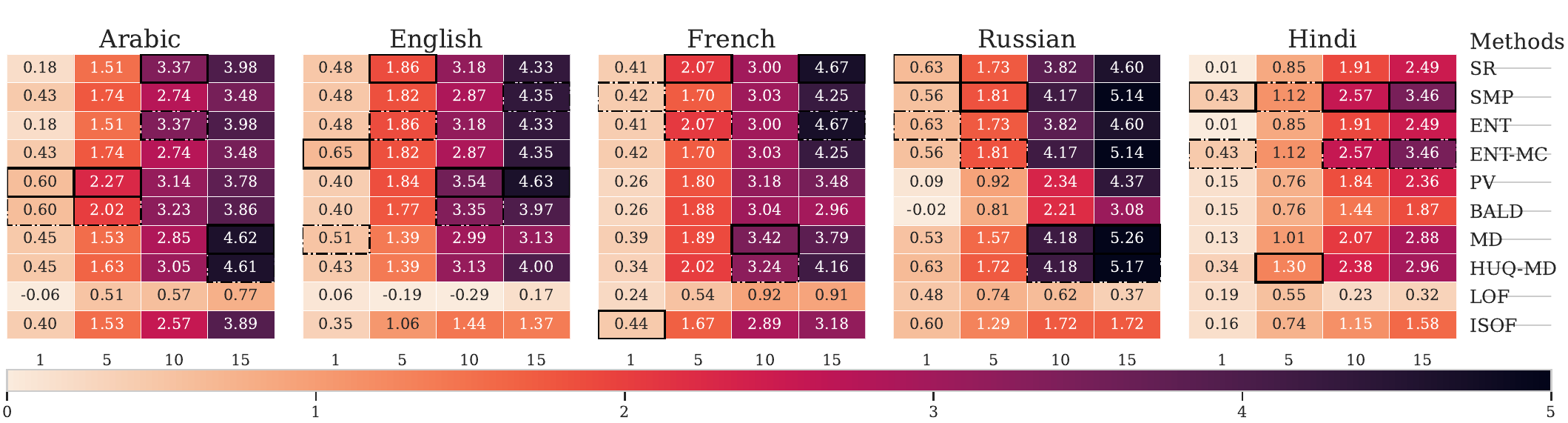}
    \subcaption{In-domain (cross-validation on the Readme dataset)}\label{fig:one:a}
  \end{subfigure}

  \vspace{4pt} 

  \begin{subfigure}{\textwidth}
    \centering
    \includegraphics[width=\linewidth,trim=2pt 2pt 2pt 2pt,clip]{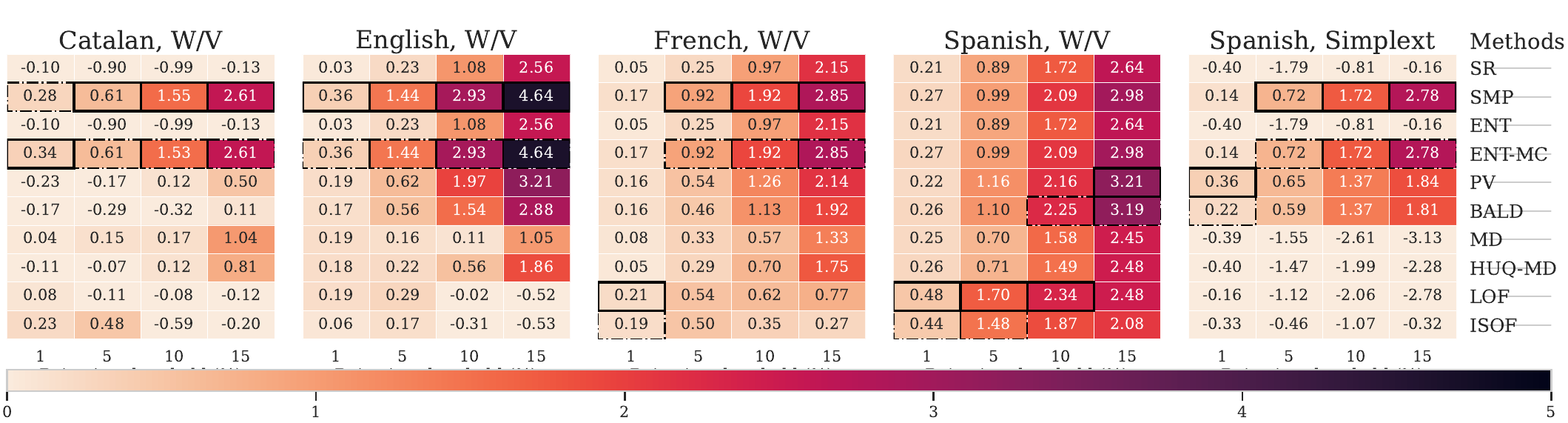}
    \subcaption{Out-of-domain (training on Readme; testing on Vikidia/Wikipedia and Simplext)}\label{fig:one:b}
  \end{subfigure}

  \caption{Improvement measured via $\Delta$F1 over the baseline from Table \ref{tab:full_metrics_with_accuracy} across rejection thresholds (Columns for 1, 5, 10, 15\%) per language per UE score.
    A solid black box marks the best UE score per column and a dash–dot box marks the second best. The UE scores are labelled on the right.}
  \label{fig:one}
\end{figure*}

We test across the five core languages in Readme as in-domain, and add two languages (Catalan, Spanish) out-of-domain on Vikidia/Wikipedia and Simplext. This setup probes robustness under both language and domain shift, with in-domain vs.~OOD results visualised in Fig.~\ref{fig:one} (panels~\ref{fig:one:a} and~\ref{fig:one:b}).  The $\Delta$F1 gain is the difference between predicting on the full test set and after abstaining for the rejected test items.

In the in-domain setting, abstention at low rejection thresholds (1–15\%) can meaningfully improve robustness, especially for the higher resource languages (the gains are a bit lower for Hindi), as using UE scores yields consistent $\Delta$F1 gains, with no method as a clear winner.  The cheap SR and ENT offer good performance, especially for English and French. However, robustness under distribution shift remains a major challenge: in the OOD setting, improvements shrink and become less stable.  Especially distance-based methods (LOF, ISOF, MD) show high variability, sometimes with negative improvements, while MC-Dropout based approaches (SMP and ENT-MC) remain comparatively strong.  

Error analysis of the most uncertain incorrect predictions (10\% threshold) shows that the UE methods detect atypical examples, for example, narrative multi-clause or punctuation-dense structures,  atypical lexical patterns, as well as shifts in domains (health/politics/tech) or genres (technical texts), which were missing from the training set.

\section{Related Work}
\label{sec:related_work}

Uncertainty estimation (UE) has become a crucial component in assessing the reliability of modern neural networks, particularly in safety-critical or noisy natural language processing (NLP) tasks. Early approaches derived uncertainty directly from model outputs, for example using the maximum \textit{softmax response (SR)} as a confidence proxy \citep{Hendrycks2017}. However, such probabilistic measures are often poorly calibrated and overconfident under dataset shift \citep{Guo2017, Liang2018}.

To better capture epistemic uncertainty, stochastic approximation techniques such as Monte Carlo Dropout (MC-Dropout) \citep{gal2016dropout} and Deep Ensembles \citep{lakshminarayanan2017} have been proposed. These approaches generate predictive distributions by repeated stochastic forward passes or model ensembles. Recent studies show that MC-Dropout-derived scores, such as \textit{Sampled Max Probability (SMP)} and entropy-based variants (\textit{ENT-MC}), achieve more reliable uncertainty discrimination and calibration than SR, especially under domain or language shift \citep{ovadia2019can}.

Complementary approaches evaluate uncertainty via feature-space geometry. Mahalanobis Distance (MD) \citep{lee2018simple} measures the distance of a representation to class centroids, while unsupervised detectors such as Local Outlier Factor (LOF) \citep{breunig2000lof} and Isolation Forest (ISOF) \citep{liu2008isolation} identify samples lying far from the training distribution. These methods often improve error detection but may exhibit instability across datasets due to sensitivity to embedding space variability.

More recent methods combine complementary uncertainty signals. The \textit{Hybrid Uncertainty Quantification} (HUQ-MD) approach \citep{vazhentsev23uncertainty} integrates probabilistic and geometric cues, balancing aleatoric and epistemic factors. Hybrid models have been shown to maintain robust calibration and selective prediction performance across multiple NLP benchmarks.

Prior work has primarily focused on English and in-domain evaluations, leaving multilingual and noisy settings underexplored \citep{Lang2023}. Moreover, UE metrics, such as calibration error (ECE), ROC-AUC, and risk–coverage AUC, often correlate weakly across datasets, complicating fair comparisons\citep{ovadia2019can}. Our study addresses these gaps by systematically evaluating nine UE methods across multiple languages and noise conditions. 


\section{Conclusions and future work}
\label{sec:conclusions}

Our results show that:
\begin{itemize}
    \item Classifier performance degrades under domain and language shifts, and the same holds for the quality of UE scoring.
    \item Stronger UE performance on in-domain data does not necessarily translate into better prediction under dataset shift.
    \item SR and ENT are accurate and computationally efficient UE measures, which are often close to the best-performing methods, which makes them a safer choice in higher-resource in-domain settings (rejecting 10\% of the lowest SR scores boosts the F1 score from 0.81 to 0.85). However, they become far less robust under domain and language shifts.
    \item MC-Dropout-based UE methods consistently exhibit better performance under both domain and language variation, particularly in lower-resource languages where softmax confidence begins to break down.
    \item Traditional outlier detection methods (such as ISOF) as well as BALD, and PV can perform well as UE scorers, often rivalling more recent approaches, but their high variance and sensitivity to data shifts mean they cannot always be trusted in practice, while getting trustable predictions is precisely the reason for using UE scores.  Also they require access to the training set to estimate the distance from it, which can be limiting in practice.
\end{itemize}

Ultimately, our findings return us to the central question of this paper: \textbf{to predict or not to predict} is less about finding a single best UE score and
more about balancing robustness and computational efficiency.

As future work, we plan to develop a meta-uncertainty framework capable of adaptively selecting or combining uncertainty metrics based on data context. By learning from fold-level and input-level features, such a meta-model could choose the most suitable metric or blend multiple scores to improve uncertainty estimation across domains and conditions. 
Future research should not only report discrimination and calibration scores, but also directly test how uncertainty estimates translate into selective prediction under both in-domain and OOD conditions.  
Only by aligning these perspectives can we design UE methods that are not merely theoretically sound but also practically reliable for real-world decision-making.
More research is also needed to determine how our findings generalise to multi-class predictions across a wider range of text classification tasks.

In a related study \cite{khallaf26noise}, we have also investigated the impact of dataset noise in a similar text classification task.  This shows that, as expected, noise in the training set reduces performance, while automatic detection of noise (for example, using Gaussian Mixture Models) can improve it.  In the next step, it is important to investigate how noise relates to uncertainty on a scale beyond manual error analysis conducted in our current study above.

\section{Acknowledgements}
This document is part of a project that has received funding from the European Union’s Horizon Europe research and innovation program under Grant Agreement No. 101132431 (iDEM Project). The University of Leeds was funded by UK Research and Innovation (UKRI) under the UK government’s Horizon Europe funding guarantee (Grant Agreement No. 10103529). The views and opinions expressed in this document are solely those of the author(s) and do not necessarily reflect the views of the European Union. Neither the European Union nor the granting authority can be held responsible for them. We are grateful to Alex Panchenko, Artem Shelmanov and Artem Vazhentsev for their comments on the earlier drafts of the paper.

\section{Bibliographical References}\label{sec:reference}

\bibliographystyle{lrec2026-natbib}

\bibliography{bibexport}
\appendix
\input{appendix}

\end{document}

%% file: appendix.tex
\onecolumn
\appendix
\section{Uncertainty Measures}
\label{app:uncertainty}
\begin{minipage}{\textwidth}

\centering
\renewcommand{\arraystretch}{1.25}
\setlength{\tabcolsep}{5pt}

\begin{ThreePartTable}

  \begin{tabularx}{\textwidth}{|l|l|L{5.2cm}|X|}
  \hline
  \textbf{Code} & \textbf{Type} & \textbf{Formula} & \textbf{Description} \\
  \hline
  \multicolumn{4}{|l|}{\textbf{Class Probability–based Methods}} \\ \hline
  SR & Aleatoric & $u_{\text{SR}} = 1 - \max_{c \in C} p^c$ \par\vspace{2pt}
  \scriptsize{where \( C \) is the number of classes, \( p_t^c \) is the predicted probability for class \( c \) during the \( t \)-th forward pass.} 
  &
      Uses the maximum softmax probability as a proxy for confidence. Higher maximum $\Rightarrow$ lower uncertainty \citep{Guo2017}. \\
  SMP & Epistemic & $u_{\text{SMP}} = 1 - \max_{c \in C}\!\big(\tfrac{1}{T}\sum_{t=1}^{T} p_t^c\big)$ \par\vspace{2pt}
  \scriptsize{where \( p_t^c \) denotes the predicted probability for class \( c \) during the \( t \)-th forward pass.}
  &
      Averages top-class probability over $T$ MC-dropout passes; captures parameter uncertainty \citep{gal2016dropout}. \\
  ENT & Aleatoric & $u_{\text{ENT}} = -\sum_{c=1}^{C} p^c \log p^c$ &
      Ambiguity of the class probability distribution \citep{ovadia2019can}. \\
  ENT-MC & Total (Predictive) & $u_{\text{ENT-MC}} = -\sum_{c=1}^{C} \overline{p^c}\log \overline{p^c}$ &
      Entropy of mean probabilities across MC passes; captures aleatoric+epistemic \citep{ovadia2019can}. \\
  PV & Epistemic & $u_{\text{PV}} = \tfrac{1}{C}\sum_{c=1}^{C}\!\big(\tfrac{1}{T}\sum_{t=1}^{T}(p_t^c-\overline{p^c})^2\big)$ \par\vspace{2pt}
  \scriptsize{where \( C \) is the number of classes, \( p_t^c \) is the predicted probability for class \( c \) during the \( t \)-th forward pass.} 
  &
      Variability in predicted probabilities across MC passes \citep{lakshminarayanan2017}. \\
  BALD & Epistemic & $u_{\text{BALD}} = u_{\text{ENT-MC}} + \tfrac{1}{T}\sum_{t=1}^{T}\sum_{c=1}^{C} p_t^c \log p_t^c$ \par\vspace{2pt}
  \scriptsize{the first term corresponds to the entropy of the mean prediction (i.e., total predictive uncertainty), while the second term represents the expected entropy over all passes (i.e., data or aleatoric uncertainty)}
  &
      Mutual information between predictions and weights; separates total vs. aleatoric parts \citep{gal2016dropout}. \\
  \hline
  \multicolumn{4}{|l|}{\textbf{Feature–based Methods}} \\ \hline
  MD & Epistemic & $ \min_{c\in C}(h_i-\mu_c)^{\!\top}\Sigma^{-1}(h_i-\mu_c)$ \par\vspace{2pt}
  \scriptsize{where \( h_i \) is the hidden representation of the \( i \)-th test instance, \( \mu_c \) is the mean embedding (centroid) for class \( c \), and \( \Sigma \) is the shared covariance matrix estimated from training data.}
  &
      Distance in latent space to class centroids; effective for OOD \citep{podolskiy2021revisiting}. \\
  LOF & Epistemic / OOD & $\frac{1}{|N_{\min}(x)|}\sum_{o\in N_{\min}(x)}\frac{lrd(o)}{lrd(x)}$ \par\vspace{2pt}
  \scriptsize{comparing their local reachability density ($lrd$) to nearby training examples $N_{Min}$}
  
  &
      Local density deviation vs. neighbors \citep{breunig2000lof}. \\
  ISOF & Epistemic / OOD & $-2^{-\tfrac{1}{N}\sum_{i=1}^{N} l_i(x)}$ \par\vspace{2pt}
  \scriptsize{where $l_i$ is the length of paths of trees in the set, see \citep{liu2008isolation} for the method of their construction. 
To match our “higher = more uncertain” convention, we invert the LOF density output and normalise the result, so larger values mean more outlier-like (and thus more uncertain).}
  &
      Isolation depth in random trees; shorter paths $\Rightarrow$ more anomalous \citep{liu2008isolation}. \\
  \hline
  \multicolumn{4}{|l|}{\textbf{Hybrid Methods}} \\ \hline
  HUQ-MD & Combined & $U_T(x)=(1-\alpha)R(U_E(x),\mathcal{D})+\alpha R(U_A(x),\mathcal{D})$\par\vspace{2pt}
  \scriptsize{where \( h_i \) is the hidden representation of the \( i \)-th test instance, \( \mu_c \) is the mean embedding (centroid) for class \( c \), and \( \Sigma \) is the shared covariance matrix estimated from training data. For consistency with other measures, we convert the model’s scores so that higher values indicate greater uncertainty, and we normalise them to a common scale.}
  &
      Rank-based mix of epistemic and aleatoric; $\alpha\!\in\![0,1]$ tunes the trade-off. \\
  \hline
  \end{tabularx}

\end{ThreePartTable}
 \captionof{table}{Summary of uncertainty estimation methods split by formula and description.
  Methods are grouped by family and labeled by the uncertainty type—\textbf{aleatoric},
  \textbf{epistemic}, and \textbf{total/predictive}.}
  \label{tab:uncertainty_summary_all}

\end{minipage}

\noindent
\par\vspace{5pt}
\begin{minipage}{\textwidth}
\footnotesize
\textbf{Notes.}
\textbf{Aleatoric:} data noise/ambiguity (e.g., overlapping classes); captured by single-pass entropy (ENT).

\textbf{Epistemic:} model uncertainty due to limited data; estimated via variability across stochastic passes (PV, BALD).

\textbf{Total (Predictive):} overall uncertainty after marginalizing model parameters (aleatoric+epistemic); entropy of averaged predictive distribution (ENT-MC).
\end{minipage}

\section{Uncertainty evaluation metrics}
\label{app:metrics_eval}

\begin{center}
\begin{minipage}{\textwidth}
\renewcommand{\arraystretch}{1.4}
\setlength{\tabcolsep}{6pt}

\begin{ThreePartTable}
  
  \begin{tabularx}{\textwidth}{|l|L{4.8cm}|X|}
  \hline
  \textbf{Code} & \textbf{Formula} & \textbf{Description} \\
  \hline

  \multicolumn{3}{|l|}{\textbf{Discrimination / Ranking}} \\ \hline

  ROC-AUC &
\makecell[tl]{\small \fiteq{\mathrm{ROC\text{-}AUC}
 = \mathrm{roc\_auc\_score}(\{y_i\},\,\{c_i\})}\\[2pt]
  \scriptsize $y_i{=}1$ if correct else $0$} &
  Ranks confidence of correct vs. incorrect predictions across thresholds. \\

  AU-PRC &
  {\small $\mathrm{AU\text{-}PRC}=\int_{0}^{1}\mathrm{Precision}(r)\,d\mathrm{Recall}(r)$} &
  Area under precision–recall curve; preferable to ROC when positives (or correctness) are imbalanced. \\

  \hline
  \multicolumn{3}{|l|}{\textbf{Calibration}} \\ \hline

  Cal. Slope &
  \makecell[tl]{\small $y_i=\alpha+\beta\,c_i+\varepsilon_i$ \\[2pt]
  \scriptsize $\beta$ by OLS} &
  Ideal $\beta{=}1$; $\beta{<}1$ over-confident, $\beta{>}1$ under-confident. \\

  CITL &
  {\small $\mathrm{CITL}=\frac{1}{n}\sum_{i=1}^n c_i - \frac{1}{n}\sum_{i=1}^n y_i$} &
  Calibration-in-the-large; $0$ indicates perfect average calibration \citep{naeini15calibration,kumar19calibration}. \\

  ECE &
  {\small $\mathrm{ECE}=\sum_{m=1}^{M}\frac{|B_m|}{n}\left|\mathrm{acc}(m)-\mathrm{conf}(m)\right|$} &
  Expected Calibration Error with $M$ bins; compares accuracy vs. mean confidence per bin \citep{ao23failure}. \\

  \hline
  \multicolumn{3}{|l|}{\textbf{Selective Prediction / Risk–Coverage}} \\ \hline

  RC-AUC &
  \makecell[tl]{\small $\mathrm{RC\text{-}AUC}=\sum_{k=1}^{n}\frac{1}{k}\sum_{i=1}^{k} r_{(i)}$ \\[2pt]
  \scriptsize $r_{(i)}$: risks ordered by decreasing confidence} &
  Area under the risk–coverage curve; lower cumulative risk at higher coverage is better. \\

  NRC-AUC &
\makecell[tl]{\small \fiteq{
\mathrm{NRC\text{-}AUC} =
\frac{\mathrm{RC\text{-}AUC}_{\text{model}} - \mathrm{RC\text{-}AUC}_{\text{random}}}
     {\mathrm{RC\text{-}AUC}_{\text{oracle}} - \mathrm{RC\text{-}AUC}_{\text{random}}}
}}&
  Normalizes RC-AUC to $[0,1]$ between random and oracle sorting; $1$ is oracle-level \citep{vazhentsev2025uncertainty}. \\

  E-AUoptRC &
  \makecell[tl]{\small $\mathrm{E\text{-}AUoptRC}=\int_{0}^{c^*} r(c)\,dc$ \\[2pt]
  \scriptsize $c^*$: coverage defined by full-set macro-F$_1$} &
  Expected area under risk–coverage up to the optimal coverage $c^*$ \citep{geifman2017selective}. \\

  \hline
  \multicolumn{3}{|l|}{\textbf{Trust / Thresholded Performance}} \\ \hline

  TI &
  \makecell[tl]{\small $\mathrm{TI}=\mathrm{F}_1(\{(\hat{y}_i,y_i)\}_{i=1}^{k})$ \\[2pt]
  \scriptsize $k=\lfloor c^* n\rfloor$} &
  F1 on the $k$ most confident predictions (coverage $c^*$); measures trustworthiness of the most confident subset \citep{ao23failure}. \\

  \hline
  \end{tabularx}

\end{ThreePartTable}

\captionof{table}{Summary of uncertainty evaluation metrics split by \emph{Formula} and \emph{Description}. Metrics are grouped into discrimination/ranking, calibration, selective prediction (risk–coverage), and trust-related measures.}
  \label{tab:uncertainty_metrics}

\end{minipage}
\end{center}
\noindent
\par\vspace{5pt}
\begin{minipage}{\textwidth}
\footnotesize
  
    \textbf{Note:} Notation: $y_i\!\in\!\{0,1\}$ (correctness), $c_i\!\in\![0,1]$ (confidence), $r(c)$ risk at coverage $c$,
    $r_{(i)}$ risks sorted by decreasing confidence, $c^*$ coverage at overall accuracy,
    $\hat{y}_i$ predicted label, $n$ dataset size, $k=\lfloor c^* n\rfloor$.
 
\end{minipage}
\begin{minipage}[t]{\textwidth}
\section*{Methods Summary}
\phantomsection
\addcontentsline{toc}{section}{Methods Summary}
\vspace*{-0.3\baselineskip}
\centering
\small                      
\setlength{\tabcolsep}{2pt}             
\renewcommand{\arraystretch}{1.19}      

\makebox[\linewidth][c]{%
\resizebox{\linewidth}{!}{%
\begin{tabular}{l|r@{\hspace{0.1cm}}r|r@{\hspace{0.1cm}}r|r@{\hspace{0.1cm}}r|r@{\hspace{0.1cm}}r|r@{\hspace{0.1cm}}r|r@{\hspace{0.1cm}}r|r@{\hspace{0.1cm}}r|r@{\hspace{0.1cm}}r|r@{\hspace{0.1cm}}r|r@{\hspace{0.1cm}}r}
\hline
\multicolumn{20}{l}{\textbf{Arabic}} \\
\hline
Metric & \multicolumn{2}{c|}{SR} & \multicolumn{2}{c|}{SMP} & \multicolumn{2}{c|}{ENT} & \multicolumn{2}{c|}{ENT-MC} & \multicolumn{2}{c|}{PV} & \multicolumn{2}{c|}{BALD} & \multicolumn{2}{c|}{MD} & \multicolumn{2}{c|}{HUQ-MD} & \multicolumn{2}{c|}{LOF} & \multicolumn{2}{c}{ISOF} \\
& $\mu$&$\sigma$&$\mu$&$\sigma$&$\mu$&$\sigma$&$\mu$&$\sigma$&$\mu$&$\sigma$&$\mu$&$\sigma$&$\mu$&$\sigma$&$\mu$&$\sigma$&$\mu$&$\sigma$&$\mu$&$\sigma$\\
\hline
ROC-AUC & 0.64 & 0.055 & 0.63 & 0.077 & 0.64 & 0.055 & 0.63 & 0.077 & 0.64 & 0.069 & 0.62 & 0.056 & 0.67 & 0.075 & 0.67 & 0.064 & 0.58 & 0.046 & \textbf{0.73} & 0.067 \\
AU-PRC & \underline{0.30} & 0.042 & \underline{0.32} & 0.035 & \underline{0.30} & 0.042 & \underline{0.32} & 0.035 & \underline{0.36} & 0.069 & \underline{0.36} & 0.059 & \underline{0.34} & 0.047 & \underline{0.34} & 0.032 & 0.22 & 0.046 & \textbf{0.38} & 0.114 \\
\hline
C-Slope & \underline{0.37} & 0.212 & 0.35 & 0.204 & 0.29 & 0.188 & 0.27 & 0.188 & \underline{0.74} & 0.282 & \textbf{0.77} & 0.279 & \underline{0.56} & 0.134 & 0.34 & 0.100 & 0.27 & 0.182 & \underline{0.50} & 0.218 \\
CITL & \textbf{-0.00} & 0.055 & \underline{-0.04} & 0.056 & -0.09 & 0.091 & \underline{-0.16} & 0.069 & 0.12 & 0.036 & 0.11 & 0.034 & \underline{0.09} & 0.058 & -0.33 & 0.036 & 0.05 & 0.025 & -0.15 & 0.064 \\
ECE & \underline{0.15} & 0.082 & \underline{0.16} & 0.074 & \underline{0.20} & 0.126 & 0.24 & 0.105 & \underline{0.14} & 0.043 & \underline{0.13} & 0.038 & 0.14 & 0.025 & 0.34 & 0.029 & \textbf{0.11} & 0.020 & \underline{0.17} & 0.054 \\
\hline
RC-AUC & 0.88 & 0.030 & 0.87 & 0.038 & 0.88 & 0.030 & 0.87 & 0.038 & 0.87 & 0.050 & 0.86 & 0.052 & 0.87 & 0.053 & 0.88 & 0.046 & \underline{0.80} & 0.038 & \textbf{0.72} & 0.063 \\
N.RC-AUC & \underline{0.35} & 0.114 & \underline{0.30} & 0.152 & \underline{0.35} & 0.114 & \underline{0.30} & 0.152 & \underline{0.28} & 0.203 & 0.23 & 0.196 & \underline{0.33} & 0.206 & \underline{0.39} & 0.180 & 0.24 & 0.141 & \textbf{0.51} & 0.113 \\
E-AUoptRC & 0.23 & 0.067 & 0.23 & 0.081 & 0.22 & 0.067 & 0.23 & 0.081 & \underline{0.17} & 0.112 & \underline{0.16} & 0.104 & 0.22 & 0.056 & \underline{0.18} & 0.063 & \textbf{0.11} & 0.020 & \underline{0.17} & 0.084 \\
TI & \underline{0.87} & 0.039 & \underline{0.87} & 0.039 & \underline{0.87} & 0.039 & \underline{0.87} & 0.039 & \underline{0.87} & 0.041 & \underline{0.87} & 0.039 & \underline{0.88} & 0.040 & \textbf{0.88} & 0.036 & 0.83 & 0.035 & \underline{0.87} & 0.055 \\
TI@95 & \underline{0.84} & 0.036 & \underline{0.85} & 0.034 & \underline{0.84} & 0.036 & \underline{0.85} & 0.034 & \textbf{0.85} & 0.038 & \underline{0.85} & 0.040 & 0.84 & 0.035 & \underline{0.84} & 0.034 & 0.83 & 0.037 & \underline{0.84} & 0.038 \\
\hline
\multicolumn{20}{l}{\textbf{English}} \\ \hline
ROC-AUC & \underline{0.70} & 0.109 & \underline{0.65} & 0.115 & \underline{0.70} & 0.109 & \underline{0.65} & 0.115 & \underline{0.63} & 0.082 & \underline{0.61} & 0.085 & \underline{0.72} & 0.051 & \textbf{0.73} & 0.047 & 0.55 & 0.037 & \underline{0.73} & 0.112 \\
AU-PRC & \textbf{0.31} & 0.060 & \underline{0.29} & 0.055 & \underline{0.31} & 0.060 & \underline{0.28} & 0.055 & \underline{0.28} & 0.045 & \underline{0.27} & 0.054 & \underline{0.29} & 0.065 & \underline{0.30} & 0.032 & 0.13 & 0.020 & \underline{0.29} & 0.129 \\
\hline
C-Slope & \underline{0.44} & 0.302 & 0.40 & 0.283 & 0.33 & 0.258 & 0.32 & 0.259 & \textbf{0.75} & 0.212 & \underline{0.72} & 0.250 & \underline{0.68} & 0.326 & 0.31 & 0.044 & 0.14 & 0.150 & 0.36 & 0.228 \\
CITL & \underline{-0.08} & 0.090 & \underline{-0.11} & 0.069 & \underline{-0.18} & 0.159 & -0.25 & 0.113 & 0.07 & 0.019 & 0.06 & 0.028 & \underline{0.05} & 0.033 & -0.39 & 0.011 & \textbf{-0.05} & 0.099 & -0.15 & 0.091 \\
ECE & \underline{0.15} & 0.096 & \underline{0.16} & 0.096 & \underline{0.24} & 0.155 & \underline{0.28} & 0.137 & \underline{0.08} & 0.021 & \textbf{0.08} & 0.017 & \underline{0.09} & 0.035 & 0.39 & 0.015 & \underline{0.10} & 0.054 & \underline{0.16} & 0.089 \\
\hline
RC-AUC & 0.92 & 0.034 & 0.91 & 0.034 & 0.92 & 0.034 & 0.91 & 0.034 & 0.91 & 0.025 & 0.91 & 0.022 & 0.93 & 0.024 & 0.94 & 0.023 & \underline{0.88} & 0.016 & \textbf{0.81} & 0.072 \\
N.RC-AUC & \underline{0.31} & 0.281 & \underline{0.22} & 0.313 & \underline{0.31} & 0.281 & \underline{0.22} & 0.313 & \underline{0.20} & 0.308 & 0.18 & 0.338 & \underline{0.41} & 0.231 & \underline{0.47} & 0.198 & 0.13 & 0.138 & \textbf{0.51} & 0.198 \\
E-AUoptRC & \underline{0.14} & 0.062 & \underline{0.14} & 0.071 & \underline{0.14} & 0.063 & \underline{0.14} & 0.071 & \underline{0.10} & 0.058 & \textbf{0.10} & 0.046 & \underline{0.23} & 0.094 & \underline{0.19} & 0.091 & \underline{0.14} & 0.053 & \underline{0.25} & 0.095 \\
TI & \underline{0.90} & 0.019 & \textbf{0.90} & 0.012 & \underline{0.90} & 0.019 & 0.90 & 0.012 & \underline{0.90} & 0.007 & \underline{0.89} & 0.012 & \underline{0.88} & 0.034 & \underline{0.89} & 0.024 & 0.86 & 0.019 & \underline{0.87} & 0.036 \\
TI@95 & \underline{0.88} & 0.022 & \underline{0.87} & 0.022 & \underline{0.88} & 0.022 & \underline{0.87} & 0.022 & \underline{0.88} & 0.018 & \textbf{0.88} & 0.016 & \underline{0.87} & 0.020 & 0.87 & 0.020 & 0.85 & 0.021 & 0.87 & 0.009 \\
\hline
\multicolumn{20}{l}{\textbf{French}} \\ \hline
ROC-AUC & \underline{0.64} & 0.141 & 0.62 & 0.123 & \underline{0.64} & 0.141 & 0.62 & 0.123 & 0.63 & 0.049 & 0.61 & 0.066 & \underline{0.75} & 0.051 & \underline{0.76} & 0.057 & 0.59 & 0.069 & \textbf{0.78} & 0.097 \\
AU-PRC & \underline{0.31} & 0.074 & 0.26 & 0.050 & \underline{0.31} & 0.073 & 0.26 & 0.050 & \underline{0.28} & 0.037 & \underline{0.27} & 0.069 & \underline{0.33} & 0.053 & \textbf{0.33} & 0.037 & 0.20 & 0.055 & \underline{0.33} & 0.106 \\
\hline
C-Slope & \underline{0.34} & 0.243 & 0.36 & 0.145 & \underline{0.24} & 0.268 & 0.27 & 0.174 & \underline{0.59} & 0.193 & 0.58 & 0.189 & \textbf{0.81} & 0.298 & \underline{0.38} & 0.088 & 0.35 & 0.209 & \underline{0.43} & 0.180 \\
CITL & \underline{-0.05} & 0.116 & \underline{-0.09} & 0.100 & \underline{-0.15} & 0.189 & -0.22 & 0.137 & \underline{0.07} & 0.029 & \underline{0.06} & 0.032 & \underline{0.07} & 0.037 & -0.38 & 0.022 & \textbf{0.01} & 0.080 & -0.13 & 0.105 \\
ECE & \underline{0.16} & 0.075 & \underline{0.17} & 0.077 & \underline{0.25} & 0.134 & \underline{0.28} & 0.119 & 0.10 & 0.035 & \textbf{0.09} & 0.036 & \underline{0.10} & 0.032 & 0.38 & 0.021 & \underline{0.10} & 0.023 & \underline{0.16} & 0.069 \\
\hline
RC-AUC & 0.91 & 0.049 & 0.90 & 0.047 & 0.91 & 0.049 & 0.90 & 0.047 & \underline{0.90} & 0.028 & \underline{0.90} & 0.026 & 0.94 & 0.023 & 0.95 & 0.021 & 0.83 & 0.048 & \textbf{0.76} & 0.051 \\
N.RC-AUC & \underline{0.26} & 0.365 & \underline{0.20} & 0.383 & \underline{0.26} & 0.365 & \underline{0.20} & 0.383 & 0.19 & 0.222 & 0.19 & 0.212 & 0.60 & 0.161 & \underline{0.63} & 0.159 & 0.25 & 0.152 & \textbf{0.67} & 0.150 \\
E-AUoptRC & \underline{0.16} & 0.105 & \underline{0.17} & 0.102 & \underline{0.16} & 0.107 & \underline{0.17} & 0.102 & \underline{0.12} & 0.031 & \underline{0.11} & 0.025 & \underline{0.15} & 0.135 & \textbf{0.08} & 0.037 & \underline{0.09} & 0.019 & \underline{0.15} & 0.116 \\
TI & \textbf{0.92} & 0.030 & \underline{0.91} & 0.031 & 0.92 & 0.030 & \underline{0.91} & 0.031 & \underline{0.90} & 0.030 & \underline{0.90} & 0.038 & \underline{0.90} & 0.048 & \underline{0.91} & 0.032 & 0.88 & 0.020 & \underline{0.90} & 0.048 \\
TI@95 & \textbf{0.89} & 0.036 & \underline{0.88} & 0.032 & 0.89 & 0.036 & \underline{0.88} & 0.032 & \underline{0.89} & 0.031 & \underline{0.89} & 0.031 & \underline{0.89} & 0.039 & \underline{0.89} & 0.038 & \underline{0.87} & 0.029 & \underline{0.88} & 0.036 \\
\hline
\multicolumn{20}{l}{\textbf{Russian}} \\ \hline
ROC-AUC & 0.65 & 0.115 & 0.64 & 0.141 & 0.65 & 0.115 & 0.64 & 0.141 & 0.64 & 0.049 & 0.62 & 0.051 & \underline{0.75} & 0.058 & \textbf{0.75} & 0.062 & 0.62 & 0.042 & \underline{0.74} & 0.101 \\
AU-PRC & \underline{0.37} & 0.148 & \underline{0.36} & 0.170 & \underline{0.37} & 0.148 & \underline{0.36} & 0.170 & \underline{0.29} & 0.080 & \underline{0.27} & 0.084 & \underline{0.39} & 0.123 & \textbf{0.40} & 0.136 & \underline{0.25} & 0.021 & \underline{0.36} & 0.140 \\
\hline
C-Slope & \underline{0.40} & 0.357 & \underline{0.42} & 0.382 & \underline{0.35} & 0.340 & \underline{0.34} & 0.332 & \underline{0.42} & 0.305 & \underline{0.38} & 0.274 & \textbf{0.52} & 0.395 & \underline{0.45} & 0.137 & \underline{0.48} & 0.152 & \underline{0.48} & 0.194 \\
CITL & \textbf{-0.02} & 0.060 & \underline{-0.05} & 0.046 & -0.12 & 0.116 & -0.19 & 0.065 & 0.11 & 0.030 & 0.10 & 0.033 & 0.09 & 0.060 & -0.33 & 0.030 & \underline{0.03} & 0.039 & \underline{-0.11} & 0.044 \\
ECE & \underline{0.15} & 0.083 & \underline{0.16} & 0.086 & \underline{0.22} & 0.129 & 0.26 & 0.115 & 0.14 & 0.028 & 0.14 & 0.029 & \underline{0.13} & 0.047 & 0.35 & 0.022 & \textbf{0.09} & 0.014 & \underline{0.12} & 0.041 \\
\hline
RC-AUC & 0.87 & 0.051 & 0.87 & 0.050 & 0.87 & 0.051 & 0.87 & 0.050 & 0.88 & 0.027 & 0.87 & 0.031 & 0.90 & 0.048 & 0.91 & 0.040 & \underline{0.78} & 0.037 & \textbf{0.73} & 0.086 \\
N.RC-AUC & \underline{0.20} & 0.393 & \underline{0.16} & 0.356 & \underline{0.20} & 0.393 & \underline{0.16} & 0.356 & 0.25 & 0.117 & 0.21 & 0.161 & \underline{0.46} & 0.221 & \underline{0.51} & 0.178 & 0.33 & 0.179 & \textbf{0.61} & 0.180 \\
E-AUoptRC & \underline{0.18} & 0.086 & \underline{0.19} & 0.102 & \underline{0.18} & 0.086 & \underline{0.19} & 0.102 & \underline{0.13} & 0.068 & \underline{0.13} & 0.065 & \underline{0.18} & 0.108 & \underline{0.15} & 0.099 & \textbf{0.12} & 0.027 & \underline{0.13} & 0.113 \\
TI & \underline{0.87} & 0.047 & \textbf{0.87} & 0.042 & \underline{0.87} & 0.047 & 0.87 & 0.042 & \underline{0.87} & 0.022 & \underline{0.85} & 0.032 & \underline{0.86} & 0.045 & \underline{0.86} & 0.040 & 0.83 & 0.045 & \underline{0.83} & 0.081 \\
TI@95 & \underline{0.83} & 0.037 & \textbf{0.83} & 0.039 & \underline{0.83} & 0.037 & 0.83 & 0.039 & \underline{0.83} & 0.032 & \underline{0.82} & 0.030 & \underline{0.83} & 0.036 & \underline{0.83} & 0.038 & \underline{0.82} & 0.037 & \underline{0.83} & 0.038 \\
\hline
\multicolumn{20}{l}{\textbf{Hindi}} \\ \hline
ROC-AUC & \underline{0.61} & 0.115 & \underline{0.62} & 0.118 & \underline{0.61} & 0.115 & \underline{0.62} & 0.118 & 0.60 & 0.099 & 0.57 & 0.117 & \underline{0.63} & 0.015 & \underline{0.65} & 0.022 & 0.54 & 0.063 & \textbf{0.67} & 0.082 \\
AU-PRC & \underline{0.30} & 0.096 & \underline{0.33} & 0.108 & \underline{0.30} & 0.096 & \underline{0.33} & 0.105 & \underline{0.29} & 0.074 & \underline{0.28} & 0.079 & \underline{0.32} & 0.100 & \textbf{0.34} & 0.123 & 0.23 & 0.041 & \underline{0.31} & 0.067 \\
\hline
C-Slope & \underline{0.19} & 0.373 & \underline{0.24} & 0.342 & \underline{0.17} & 0.341 & \underline{0.22} & 0.324 & \underline{0.39} & 0.444 & \underline{0.29} & 0.561 & \textbf{0.41} & 0.400 & \underline{0.30} & 0.082 & \underline{0.24} & 0.255 & \underline{0.31} & 0.120 \\
CITL & \textbf{-0.01} & 0.038 & \underline{-0.04} & 0.062 & -0.11 & 0.093 & -0.18 & 0.081 & 0.14 & 0.051 & 0.13 & 0.046 & 0.10 & 0.082 & -0.31 & 0.045 & 0.07 & 0.040 & \underline{-0.09} & 0.117 \\
ECE & \underline{0.16} & 0.068 & \underline{0.16} & 0.080 & \underline{0.23} & 0.107 & \underline{0.26} & 0.106 & 0.17 & 0.044 & \underline{0.16} & 0.045 & \underline{0.15} & 0.043 & 0.34 & 0.044 & \textbf{0.12} & 0.041 & \underline{0.16} & 0.073 \\
\hline
RC-AUC & 0.84 & 0.059 & 0.85 & 0.064 & 0.84 & 0.059 & 0.85 & 0.064 & 0.84 & 0.075 & \underline{0.82} & 0.086 & 0.84 & 0.045 & 0.86 & 0.034 & \underline{0.79} & 0.038 & \textbf{0.73} & 0.021 \\
N.RC-AUC & \underline{0.23} & 0.288 & \underline{0.26} & 0.369 & \underline{0.23} & 0.288 & \underline{0.26} & 0.368 & 0.22 & 0.255 & 0.14 & 0.256 & \underline{0.20} & 0.088 & \underline{0.31} & 0.049 & 0.12 & 0.121 & \textbf{0.44} & 0.213 \\
E-AUoptRC & \underline{0.21} & 0.094 & \underline{0.20} & 0.100 & \underline{0.21} & 0.094 & \underline{0.20} & 0.100 & \underline{0.16} & 0.099 & \underline{0.17} & 0.102 & 0.28 & 0.090 & \underline{0.23} & 0.075 & \textbf{0.14} & 0.039 & 0.23 & 0.078 \\
TI & \underline{0.83} & 0.058 & \textbf{0.84} & 0.057 & \underline{0.83} & 0.058 & 0.84 & 0.057 & \underline{0.83} & 0.074 & \underline{0.83} & 0.083 & \underline{0.84} & 0.044 & \underline{0.84} & 0.041 & 0.81 & 0.065 & \underline{0.81} & 0.099 \\
TI@95 & \underline{0.81} & 0.057 & \underline{0.82} & 0.051 & \underline{0.81} & 0.057 & \underline{0.82} & 0.051 & \underline{0.81} & 0.063 & \underline{0.81} & 0.063 & \underline{0.81} & 0.049 & \textbf{0.82} & 0.046 & \underline{0.81} & 0.056 & \underline{0.81} & 0.064 \\
\bottomrule
\end{tabular}
}}
\captionof{table}{Summary of the mean ($\mu$) and standard deviation ($\sigma$) of each metric across folds for all methods grouped by the evaluation perspective. Bold indicates the best method; underlined methods are not significantly different from the best at $p=0.05$ (paired tests).}
\label{tab:metrics_summary}
\end{minipage}
\section{\texorpdfstring{$\Delta$F1}{Delta F1} across rejection thresholds}
\begin{minipage}{\textwidth}
\centering
\scriptsize
\setlength{\tabcolsep}{2pt}
\renewcommand{\arraystretch}{0.8}
\resizebox{\textwidth}{!}{%
\begin{tabular}{l *{16}{r}}
\toprule
 & \multicolumn{4}{c}{1\%} & \multicolumn{4}{c}{5\%} & \multicolumn{4}{c}{10\%} & \multicolumn{4}{c}{15\%} \\
\multirow{2}{*}{Method} & \multicolumn{2}{c}{MacroF1} & \multicolumn{2}{c}{Percent} & \multicolumn{2}{c}{MacroF1} & \multicolumn{2}{c}{Percent} & \multicolumn{2}{c}{MacroF1} & \multicolumn{2}{c}{Percent} & \multicolumn{2}{c}{MacroF1} & \multicolumn{2}{c}{Percent} \\
 &$\mu$&$\sigma$&$\mu$&$\sigma$&$\mu$&$\sigma$&$\mu$&$\sigma$&$\mu$&$\sigma$&$\mu$&$\sigma$&$\mu$&$\sigma$&$\mu$&$\sigma$\\
\midrule
\multicolumn{16}{l}{\textbf{Arabic}}\\
\midrule
SR & 0.18 & 0.41 & 33.3 & 29.8 & 1.51 & 0.49 & 44.5 & 9.1 & \textbf{3.37} & 0.44 & \textbf{46.2} & 4.4 & 3.98 & 0.97 & 38.4 & 4.0 \\
SMP & 0.43 & 0.19 & 53.3 & 16.3 & 1.74 & 0.36 & 47.7 & 6.9 & 2.74 & 0.71 & 40.1 & 2.9 & 3.48 & 1.49 & 35.4 & 6.8 \\
ENT & 0.18 & 0.40 & 33.3 & 29.8 & 1.51 & 0.49 & 44.5 & 9.1 & \underline{3.37} & 0.44 & \underline{46.2} & 4.4 & 3.98 & 0.97 & 38.4 & 4.0 \\
ENT-MC & 0.43 & 0.19 & 53.3 & 16.3 & 1.74 & 0.36 & 47.7 & 6.9 & 2.74 & 0.71 & 40.1 & 2.9 & 3.48 & 1.49 & 35.4 & 6.8 \\
PV & \textbf{0.60} & 0.37 & \textbf{66.7} & 29.8 & \textbf{2.27} & 0.47 & \textbf{58.2} & 7.3 & 3.14 & 1.40 & 44.7 & 11.4 & 3.78 & 1.54 & 37.8 & 7.1 \\
BALD & \underline{0.60} & 0.37 & \underline{66.7} & 29.8 & 2.02 & 0.64 & 53.6 & 9.2 & 3.23 & 1.12 & \underline{45.4} & 9.6 & 3.86 & 1.66 & 38.3 & 7.9 \\
MD & 0.45 & 0.18 & 53.3 & 16.3 & 1.53 & 0.19 & 44.7 & 3.9 & 2.85 & 1.08 & 41.7 & 10.3 & \textbf{4.62} & 1.71 & \textbf{42.4} & 9.4 \\
HUQ-MD & 0.45 & 0.20 & 53.3 & 16.3 & 1.63 & 0.40 & 46.0 & 7.1 & 3.05 & 0.84 & 43.2 & 8.3 & \underline{4.61} & 1.02 & \underline{41.9} & 5.5 \\
LOF & -0.06 & 0.34 & 13.3 & 26.7 & 0.51 & 0.57 & 27.1 & 9.7 & 0.57 & 1.02 & 23.6 & 8.4 & 0.77 & 1.40 & 22.3 & 8.8 \\
ISOF & 0.40 & 0.37 & 53.3 & 26.7 & 1.53 & 1.21 & 48.0 & 18.0 & 2.57 & 2.56 & 42.5 & 17.5 & 3.89 & 3.80 & 41.3 & 15.9 \\
\midrule
\multicolumn{16}{l}{\textbf{English}}\\
\midrule
SR & 0.48 & 0.20 & 55.0 & 10.0 & \textbf{1.86} & 0.45 & \textbf{43.8} & 5.0 & 3.18 & 0.31 & 37.4 & 6.6 & 4.33 & 0.39 & \underline{32.5} & 5.5 \\
SMP & 0.48 & 0.50 & 50.0 & 27.4 & \underline{1.82} & 0.69 & 40.5 & 8.7 & 2.87 & 0.51 & 34.0 & 8.6 & 4.35 & 1.07 & 31.7 & 9.5 \\
ENT & 0.48 & 0.20 & 55.0 & 10.0 & \underline{1.86} & 0.45 & \underline{43.8} & 5.0 & 3.18 & 0.31 & 37.4 & 6.6 & 4.33 & 0.39 & \underline{32.5} & 5.5 \\
ENT-MC & \textbf{0.65} & 0.39 & \textbf{60.0} & 25.5 & \underline{1.82} & 0.69 & 40.5 & 8.7 & 2.87 & 0.51 & 34.0 & 8.6 & 4.35 & 1.07 & 31.7 & 9.5 \\
PV & 0.40 & 0.50 & 45.0 & 29.2 & \underline{1.84} & 0.67 & 41.7 & 9.7 & \textbf{3.54} & 0.81 & \textbf{38.9} & 9.0 & \textbf{4.63} & 1.07 & \textbf{33.0} & 8.5 \\
BALD & 0.40 & 0.50 & 45.0 & 29.2 & \underline{1.77} & 0.53 & 41.5 & 7.4 & 3.35 & 1.06 & 37.6 & 10.4 & 3.97 & 1.64 & 29.9 & 9.8 \\
MD & 0.51 & 0.27 & 50.0 & 22.4 & 1.39 & 0.64 & 38.8 & 10.5 & 2.99 & 0.76 & 37.8 & 8.4 & 3.13 & 1.41 & 32.0 & 5.0 \\
HUQ-MD & 0.43 & 0.24 & 45.0 & 18.7 & 1.39 & 0.60 & 37.7 & 8.6 & 3.13 & 0.64 & 37.4 & 8.2 & 4.00 & 0.57 & \underline{32.5} & 4.2 \\
LOF & 0.06 & 0.30 & 25.0 & 15.8 & -0.19 & 0.50 & 15.8 & 7.1 & -0.29 & 0.51 & 14.2 & 2.7 & 0.17 & 0.51 & 15.1 & 4.2 \\
ISOF & 0.35 & 0.45 & 45.0 & 29.2 & 1.06 & 1.22 & 37.1 & 17.0 & 1.44 & 2.66 & 30.4 & 16.1 & 1.37 & 4.08 & 27.3 & 13.8 \\
\midrule
\multicolumn{16}{l}{\textbf{French}}\\
\midrule
SR & \underline{0.41} & 0.48 & \textbf{46.7} & 26.7 & \textbf{2.07} & 0.91 & \textbf{46.2} & 11.9 & 3.00 & 1.15 & 35.3 & 8.2 & \textbf{4.67} & 2.00 & \textbf{33.8} & 9.9 \\
SMP & \underline{0.42} & 0.37 & 40.0 & 24.9 & 1.70 & 0.49 & 40.9 & 4.2 & 3.03 & 0.74 & 36.6 & 2.2 & 4.25 & 1.32 & 32.4 & 6.4 \\
ENT & \underline{0.41} & 0.48 & \underline{46.7} & 26.7 & \underline{2.07} & 0.91 & \underline{46.2} & 11.9 & 3.00 & 1.15 & 35.3 & 8.2 & \underline{4.67} & 2.00 & \underline{33.8} & 9.9 \\
ENT-MC & \underline{0.42} & 0.37 & 40.0 & 24.9 & 1.70 & 0.49 & 40.9 & 4.2 & 3.03 & 0.74 & 36.6 & 2.2 & 4.25 & 1.32 & 32.4 & 6.4 \\
PV & 0.26 & 0.36 & 33.3 & 21.1 & 1.80 & 0.55 & 42.6 & 7.7 & 3.18 & 0.83 & 38.3 & 4.0 & 3.48 & 0.81 & 29.0 & 4.0 \\
BALD & 0.26 & 0.36 & 33.3 & 21.1 & 1.88 & 0.59 & 44.1 & 8.9 & 3.04 & 1.24 & 36.6 & 9.1 & 2.96 & 1.11 & 26.2 & 4.9 \\
MD & \underline{0.39} & 0.58 & \underline{46.7} & 34.0 & 1.89 & 1.19 & 42.9 & 17.4 & \textbf{3.42} & 1.38 & \textbf{40.1} & 6.7 & 3.79 & 1.77 & \underline{33.5} & 4.0 \\
HUQ-MD & 0.34 & 0.51 & 40.0 & 24.9 & \underline{2.02} & 1.13 & 44.6 & 16.0 & 3.24 & 1.75 & 37.8 & 11.3 & 4.16 & 1.37 & \underline{33.6} & 4.6 \\
LOF & 0.24 & 0.05 & 33.3 & 0.0 & 0.54 & 0.94 & 25.8 & 15.3 & 0.92 & 1.39 & 21.4 & 11.7 & 0.91 & 1.35 & 18.4 & 8.6 \\
ISOF & \textbf{0.44} & 0.43 & \underline{46.7} & 26.7 & 1.67 & 1.34 & 42.3 & 18.1 & 2.89 & 2.40 & 37.7 & 13.4 & 3.18 & 2.94 & 32.0 & 8.2 \\
\midrule
\multicolumn{16}{l}{\textbf{Russian}}\\
\midrule
SR & \textbf{0.63} & 0.53 & \textbf{63.3} & 37.1 & \underline{1.73} & 0.75 & \textbf{49.5} & 13.2 & 3.82 & 2.19 & 49.3 & 16.9 & 4.60 & 2.26 & 40.9 & 12.7 \\
SMP & \underline{0.56} & 0.48 & 56.7 & 38.9 & \textbf{1.81} & 1.33 & \underline{49.1} & 21.6 & \underline{4.17} & 2.14 & 50.3 & 17.6 & 5.14 & 2.59 & 42.8 & 14.0 \\
ENT & \underline{0.63} & 0.53 & \underline{63.3} & 37.1 & \underline{1.73} & 0.75 & \underline{49.5} & 13.2 & 3.82 & 2.19 & 49.3 & 16.9 & 4.60 & 2.26 & 40.9 & 12.7 \\
ENT-MC & \underline{0.56} & 0.48 & 56.7 & 38.9 & \underline{1.81} & 1.33 & \underline{49.1} & 21.6 & \underline{4.17} & 2.14 & 50.3 & 17.6 & 5.14 & 2.59 & 42.8 & 14.0 \\
PV & 0.09 & 0.17 & 23.3 & 20.0 & 0.92 & 0.83 & 36.2 & 14.5 & 2.34 & 1.72 & 41.4 & 14.0 & 4.37 & 1.89 & 42.5 & 11.2 \\
BALD & -0.02 & 0.25 & 13.3 & 16.3 & 0.81 & 0.91 & 34.5 & 14.4 & 2.21 & 1.88 & 40.5 & 14.7 & 3.08 & 2.71 & 36.1 & 14.1 \\
MD & \underline{0.53} & 0.43 & 56.7 & 32.7 & 1.57 & 1.26 & 45.5 & 21.8 & \textbf{4.18} & 1.77 & \textbf{51.5} & 14.3 & \textbf{5.26} & 1.68 & \textbf{44.9} & 9.5 \\
HUQ-MD & \underline{0.63} & 0.53 & \underline{63.3} & 37.1 & \underline{1.72} & 1.27 & 47.3 & 19.8 & \underline{4.18} & 1.45 & \underline{50.5} & 12.4 & \underline{5.17} & 1.50 & 43.6 & 9.9 \\
LOF & 0.48 & 0.43 & 56.7 & 22.6 & 0.74 & 0.58 & 32.4 & 8.4 & 0.62 & 1.38 & 26.7 & 7.3 & 0.37 & 1.48 & 22.6 & 4.5 \\
ISOF & \underline{0.60} & 0.53 & \underline{63.3} & 37.1 & 1.29 & 1.25 & 46.4 & 19.3 & 1.72 & 2.56 & 38.5 & 16.3 & 1.72 & 4.50 & 35.8 & 14.1 \\
\midrule
\multicolumn{16}{l}{\textbf{Hindi}}\\
\midrule
SR & 0.01 & 0.54 & 30.0 & 40.0 & 0.85 & 1.02 & 38.4 & 15.8 & 1.91 & 1.24 & 37.8 & 10.7 & 2.49 & 1.01 & 34.7 & 7.3 \\
SMP & \textbf{0.43} & 0.34 & \textbf{53.3} & 32.3 & 1.12 & 1.01 & 40.0 & 20.8 & \textbf{2.57} & 1.19 & \textbf{42.7} & 12.2 & \textbf{3.46} & 1.33 & \textbf{39.2} & 9.2 \\
ENT & 0.01 & 0.54 & 30.0 & 40.0 & 0.85 & 1.02 & 38.4 & 15.8 & 1.91 & 1.24 & 37.8 & 10.7 & 2.49 & 1.01 & 34.7 & 7.3 \\
ENT-MC & \underline{0.43} & 0.34 & \underline{53.3} & 32.3 & 1.12 & 1.01 & 40.0 & 20.8 & \underline{2.57} & 1.19 & \underline{42.7} & 12.2 & \underline{3.46} & 1.33 & \underline{39.2} & 9.2 \\
PV & 0.15 & 0.52 & 40.0 & 37.4 & 0.76 & 0.99 & 36.5 & 11.3 & 1.84 & 1.57 & 37.9 & 7.0 & 2.36 & 2.44 & 34.3 & 7.5 \\
BALD & 0.15 & 0.52 & 40.0 & 37.4 & 0.76 & 0.99 & 36.5 & 11.3 & 1.44 & 1.85 & 34.0 & 10.9 & 1.87 & 3.07 & 30.4 & 11.6 \\
MD & 0.13 & 0.42 & 36.7 & 37.1 & 1.01 & 0.78 & 40.0 & 16.5 & 2.07 & 1.06 & 39.6 & 10.5 & 2.88 & 1.24 & 37.3 & 9.3 \\
HUQ-MD & \underline{0.34} & 0.58 & 50.0 & 44.7 & \textbf{1.30} & 1.16 & \textbf{43.7} & 21.7 & 2.38 & 1.41 & 40.5 & 14.5 & 2.96 & 1.42 & 37.3 & 10.2 \\
LOF & 0.19 & 0.39 & 33.3 & 27.9 & 0.55 & 0.61 & 29.7 & 7.8 & 0.23 & 1.01 & 21.3 & 5.2 & 0.32 & 1.59 & 21.9 & 4.3 \\
ISOF & 0.16 & 0.25 & 36.7 & 19.4 & 0.74 & 1.18 & 38.5 & 14.0 & 1.15 & 2.17 & 33.9 & 12.5 & 1.58 & 3.11 & 33.0 & 8.9 \\
\bottomrule
\end{tabular}
}
\captionof{table}{Change in Macro ($\Delta$F1, in \%) after rejecting the most uncertain samples at each threshold, and the percentage of rejected incorrect predictions. Values are mean ($\mu$) and std ($\sigma$) over 5 folds.}
\label{tab:f1_improvement}
\end{minipage}
\begin{minipage}{\textwidth}
\centering
\scriptsize
\setlength{\tabcolsep}{2pt}
\renewcommand{\arraystretch}{0.9}

\centering
\resizebox{\textwidth}{!}{%
\scriptsize
\begin{tabular}{l 
@{\hspace{0.3cm}}r@{\hspace{0.1cm}}l@{\hspace{0.2cm}}r@{\hspace{0.1cm}}r @{\hspace{0.3cm}}r@{\hspace{0.1cm}}r@{\hspace{0.2cm}}r@{\hspace{0.1cm}}r @{\hspace{0.3cm}}r@{\hspace{0.1cm}}r@{\hspace{0.2cm}}r@{\hspace{0.1cm}}r @{\hspace{0.3cm}}r@{\hspace{0.1cm}}r@{\hspace{0.2cm}}r@{\hspace{0.1cm}}r}
\toprule
 & \multicolumn{4}{c}{1\%} & \multicolumn{4}{c}{5\%} & \multicolumn{4}{c}{10\%} & \multicolumn{4}{c}{15\%} \\
\multirow{2}{*}{Method} & \multicolumn{2}{c}{Macro} & \multicolumn{2}{c}{Percent} & \multicolumn{2}{c}{Macro} & \multicolumn{2}{c}{Percent} & \multicolumn{2}{c}{Macro} & \multicolumn{2}{c}{Percent} & \multicolumn{2}{c}{Macro} & \multicolumn{2}{c}{Percent} \\
 &$\mu$&$\sigma$&$\mu$&$\sigma$&$\mu$&$\sigma$&$\mu$&$\sigma$&$\mu$&$\sigma$&$\mu$&$\sigma$&$\mu$&$\sigma$&$\mu$&$\sigma$\\
\midrule
\multicolumn{16}{l}{\textbf{Catalan, Wikipedia/Vikidia}}\\
\midrule
 SR & -0.10 & 0.45 & 40.0 & 25.5 & -0.90 & 0.75 & 32.2 & 6.5 & -0.99 & 1.74 & 38.3 & 6.9 & -0.13 & 1.98 & 44.2 & 2.7 \\
 SMP & \underline{0.28} & 0.34 & 65.0 & 25.5 & \textbf{0.61} & 1.01 & \textbf{52.2} & 13.9 & \textbf{1.55} & 0.95 & \textbf{53.7} & 5.5 & \textbf{2.61} & 0.75 & \textbf{54.2} & 4.1 \\
 ENT & -0.10 & 0.45 & 40.0 & 25.5 & -0.90 & 0.75 & 32.2 & 6.5 & -0.99 & 1.74 & 38.3 & 6.9 & -0.13 & 1.98 & 44.2 & 2.7 \\
 ENT-MC & \textbf{0.34} & 0.39 & \textbf{70.0} & 29.2 & \underline{0.61} & 1.01 & \underline{52.2} & 13.9 & \underline{1.53} & 0.94 & \underline{53.7} & 5.5 & \underline{2.61} & 0.75 & \underline{54.2} & 4.1 \\
 PV & -0.23 & 0.24 & 25.0 & 15.8 & -0.17 & 1.07 & 40.0 & 8.2 & 0.12 & 1.38 & 42.3 & 6.6 & 0.50 & 2.46 & 44.6 & 6.6 \\
 BALD & -0.17 & 0.23 & 30.0 & 10.0 & -0.29 & 1.00 & 35.6 & 10.3 & -0.32 & 1.81 & 36.6 & 12.8 & 0.11 & 3.11 & 40.4 & 13.0 \\
 MD & 0.04 & 0.42 & 45.0 & 33.2 & 0.15 & 0.93 & 44.4 & 14.9 & 0.17 & 0.95 & 41.7 & 9.7 & 1.04 & 1.24 & 45.4 & 7.3 \\
 HUQ-MD & -0.11 & 0.31 & 35.0 & 25.5 & -0.07 & 0.75 & 41.1 & 13.4 & 0.12 & 0.78 & 41.7 & 6.9 & 0.81 & 1.30 & 45.0 & 6.8 \\
 LOF & 0.08 & 0.12 & 45.0 & 18.7 & -0.11 & 0.57 & 37.8 & 11.3 & -0.08 & 0.39 & 38.3 & 8.2 & -0.12 & 0.53 & 37.7 & 8.3 \\
 ISOF & \underline{0.23} & 0.28 & 60.0 & 20.0 & \underline{0.48} & 1.21 & \underline{52.2} & 15.2 & -0.59 & 2.62 & 41.1 & 12.7 & -0.20 & 4.01 & 43.5 & 13.4 \\
 
\midrule

\multicolumn{16}{l}{\textbf{English, Wikipedia/Vikidia}}\\
\midrule
 SR & 0.03 & 0.13 & 35.0 & 11.7 & 0.23 & 0.64 & 36.4 & 10.2 & 1.08 & 0.88 & 40.6 & 7.4 & 2.56 & 0.97 & 43.6 & 5.7 \\
 SMP & \textbf{0.36} & 0.24 & \textbf{61.7} & 22.4 & \textbf{1.44} & 0.36 & \textbf{54.4} & 6.4 & \textbf{2.93} & 0.54 & \textbf{53.6} & 5.5 & \textbf{4.64} & 0.94 & \textbf{53.3} & 6.2 \\
 ENT & 0.03 & 0.13 & 35.0 & 11.7 & 0.23 & 0.64 & 36.4 & 10.2 & 1.08 & 0.88 & 40.6 & 7.4 & 2.56 & 0.97 & 43.6 & 5.7 \\
 ENT-MC & \underline{0.36} & 0.24 & \underline{61.7} & 22.4 & \underline{1.44} & 0.35 & \underline{54.4} & 6.4 & \underline{2.93} & 0.54 & \underline{53.6} & 5.5 & \underline{4.64} & 0.94 & \underline{53.3} & 6.2 \\
 PV & \underline{0.19} & 0.30 & \underline{48.3} & 28.3 & 0.62 & 0.61 & 41.8 & 11.5 & 1.97 & 0.98 & 47.2 & 8.7 & 3.21 & 0.93 & 47.5 & 5.2 \\
 BALD & \underline{0.17} & 0.29 & \underline{46.7} & 27.7 & 0.56 & 0.60 & 41.1 & 11.4 & 1.54 & 0.99 & 44.2 & 8.7 & 2.88 & 1.12 & 46.1 & 6.1 \\
 MD & \underline{0.19} & 0.26 & \underline{48.3} & 22.9 & 0.16 & 0.65 & 35.0 & 10.6 & 0.11 & 0.75 & 33.2 & 5.9 & 1.05 & 1.05 & 37.2 & 5.4 \\
 HUQ-MD & 0.18 & 0.23 & \underline{48.3} & 20.3 & 0.22 & 0.54 & 36.1 & 8.9 & 0.56 & 0.80 & 36.5 & 6.7 & 1.86 & 1.12 & 40.3 & 6.3 \\
 LOF & 0.19 & 0.15 & \underline{48.3} & 13.8 & 0.29 & 0.26 & 35.5 & 5.5 & -0.02 & 0.50 & 30.2 & 4.4 & -0.52 & 0.88 & 27.6 & 4.6 \\
 ISOF & 0.06 & 0.19 & \underline{38.3} & 16.7 & 0.17 & 0.38 & 35.8 & 6.5 & -0.31 & 0.71 & 31.0 & 4.9 & -0.53 & 0.68 & 31.4 & 3.0 \\

\midrule
\multicolumn{16}{l}{\textbf{French, Wikipedia/Vikidia}}\\
\midrule
 SR & 0.05 & 0.05 & 41.6 & 4.3 & 0.25 & 0.15 & 41.4 & 2.8 & 0.97 & 0.21 & 44.2 & 1.8 & 2.15 & 0.26 & 46.4 & 1.7 \\
 SMP & 0.17 & 0.04 & 49.4 & 4.4 & \textbf{0.92} & 0.15 & \textbf{50.9} & 3.0 & \textbf{1.92} & 0.18 & \textbf{50.5} & 1.8 & \textbf{2.85} & 0.29 & \textbf{49.3} & 1.9 \\
 ENT & 0.05 & 0.04 & 41.2 & 4.2 & 0.25 & 0.15 & 41.4 & 2.8 & 0.97 & 0.21 & 44.2 & 1.8 & 2.15 & 0.26 & 46.4 & 1.7 \\
 ENT-MC & 0.17 & 0.04 & 49.4 & 4.4 & \underline{0.92} & 0.15 & \underline{50.9} & 3.0 & \underline{1.92} & 0.18 & \underline{50.5} & 1.8 & \underline{2.85} & 0.29 & \underline{49.3} & 1.9 \\
 PV & 0.16 & 0.05 & 50.5 & 5.3 & 0.54 & 0.14 & 45.6 & 2.6 & 1.26 & 0.25 & 46.6 & 2.2 & 2.14 & 0.30 & 47.1 & 1.8 \\
 BALD & 0.16 & 0.05 & 50.6 & 4.9 & 0.46 & 0.16 & 44.4 & 3.0 & 1.13 & 0.22 & 45.9 & 2.0 & 1.92 & 0.35 & 46.3 & 2.0 \\
 MD & 0.08 & 0.05 & 43.7 & 4.8 & 0.33 & 0.14 & 43.0 & 2.8 & 0.57 & 0.20 & 41.7 & 1.8 & 1.33 & 0.31 & 43.0 & 1.7 \\
 HUQ-MD & 0.05 & 0.06 & 41.6 & 5.2 & 0.29 & 0.18 & 42.2 & 3.4 & 0.70 & 0.21 & 42.4 & 1.8 & 1.75 & 0.28 & 44.6 & 1.7 \\
 LOF & \textbf{0.21} & 0.04 & \textbf{56.0} & 4.0 & 0.54 & 0.12 & 45.1 & 2.1 & 0.62 & 0.19 & 40.4 & 1.7 & 0.77 & 0.24 & 39.2 & 1.2 \\
 ISOF & \underline{0.19} & 0.06 & \underline{53.9} & 5.5 & 0.50 & 0.11 & 46.0 & 2.2 & 0.35 & 0.15 & 40.6 & 1.1 & 0.27 & 0.24 & 39.3 & 1.3 \\ 

\midrule
\multicolumn{16}{l}{\textbf{Spanish, Wikipedia/Vikidia}}\\
\midrule
 SR & 0.21 & 0.10 & 49.4 & 10.3 & 0.89 & 0.27 & 46.9 & 5.7 & 1.72 & 0.55 & 45.7 & 5.1 & 2.64 & 0.65 & 45.4 & 3.6 \\
 SMP & 0.27 & 0.15 & 56.5 & 13.2 & 0.99 & 0.40 & 49.6 & 7.4 & 2.09 & 0.51 & 49.8 & 4.9 & 2.98 & 0.56 & \underline{48.1} & 2.9 \\
 ENT & 0.21 & 0.10 & 49.4 & 10.3 & 0.89 & 0.27 & 46.9 & 5.7 & 1.72 & 0.55 & 45.7 & 5.1 & 2.64 & 0.65 & 45.4 & 3.6 \\
 ENT-MC & 0.27 & 0.15 & 56.5 & 13.2 & 0.99 & 0.40 & 49.6 & 7.4 & 2.09 & 0.51 & 49.8 & 4.9 & 2.98 & 0.56 & \underline{48.1} & 2.9 \\
 PV & 0.22 & 0.10 & 50.6 & 9.6 & 1.16 & 0.28 & 52.1 & 5.8 & 2.16 & 0.36 & 49.6 & 3.8 & \textbf{3.21} & 0.50 & \textbf{48.5} & 3.1 \\
 BALD & 0.26 & 0.10 & 54.1 & 9.4 & 1.10 & 0.29 & 50.9 & 5.9 & \underline{2.25} & 0.45 & \underline{50.4} & 4.8 & \underline{3.19} & 0.56 & \underline{48.3} & 3.2 \\
 MD & 0.25 & 0.14 & 52.9 & 11.8 & 0.70 & 0.24 & 43.2 & 5.4 & 1.58 & 0.16 & 44.4 & 1.8 & 2.45 & 0.30 & 44.2 & 2.1 \\
 HUQ-MD & 0.26 & 0.07 & 54.1 & 5.8 & 0.71 & 0.18 & 43.5 & 4.1 & 1.49 & 0.37 & 43.6 & 3.2 & 2.48 & 0.46 & 44.4 & 2.6 \\
 LOF & \textbf{0.48} & 0.13 & \textbf{75.3} & 12.0 & \textbf{1.70} & 0.25 & \textbf{62.5} & 5.3 & \textbf{2.34} & 0.20 & \textbf{51.4} & 2.3 & 2.48 & 0.15 & 44.4 & 1.6 \\
 ISOF & \underline{0.44} & 0.11 & 71.8 & 10.8 & 1.48 & 0.22 & 58.0 & 4.5 & 1.87 & 0.38 & 47.0 & 4.2 & 2.08 & 0.48 & 42.4 & 3.4 \\

\midrule
\multicolumn{16}{l}{\textbf{Spanish, Simplext}}\\
\midrule
 SR & -0.40 & 0.50 & 15.0 & 20.0 & -1.79 & 1.38 & 15.3 & 10.3 & -0.81 & 2.62 & 36.5 & 9.9 & -0.16 & 6.21 & 43.9 & 12.8 \\
 SMP & 0.14 & 0.57 & 50.0 & 35.4 & \textbf{0.72} & 2.47 & \textbf{52.9} & 30.0 & \textbf{1.72} & 4.06 & \textbf{52.9} & 22.7 & \textbf{2.78} & 6.04 & \textbf{52.9} & 19.8 \\
 ENT & -0.40 & 0.50 & 15.0 & 20.0 & -1.79 & 1.38 & 15.3 & 10.3 & -0.81 & 2.62 & 36.5 & 9.9 & -0.16 & 6.21 & 43.9 & 12.8 \\
 ENT-MC & 0.14 & 0.57 & 50.0 & 35.4 & \underline{0.72} & 2.47 & \underline{52.9} & 30.0 & \underline{1.72} & 4.06 & \underline{52.9} & 22.7 & \underline{2.78} & 6.04 & \underline{52.9} & 19.8 \\
 PV & \textbf{0.36} & 0.55 & \textbf{65.0} & 30.0 & \underline{0.65} & 1.84 & 51.8 & 12.0 & 1.37 & 3.10 & 48.8 & 14.0 & 1.84 & 4.17 & 46.3 & 12.0 \\
 BALD & \underline{0.22} & 0.50 & 55.0 & 29.2 & \underline{0.59} & 1.75 & 50.6 & 10.9 & 1.37 & 2.94 & 48.2 & 13.9 & 1.81 & 4.08 & 44.7 & 11.8 \\
 MD & -0.39 & 0.49 & 15.0 & 20.0 & -1.55 & 1.82 & 18.8 & 19.5 & -2.61 & 5.20 & 27.6 & 15.8 & -3.13 & 8.01 & 34.1 & 11.7 \\
 HUQ-MD & -0.40 & 0.50 & 15.0 & 20.0 & -1.47 & 1.93 & 20.0 & 21.6 & -1.99 & 4.15 & 30.6 & 12.0 & -2.28 & 8.26 & 37.6 & 13.5 \\
 LOF & -0.16 & 0.41 & 20.0 & 29.2 & -1.12 & 0.90 & 12.9 & 12.6 & -2.06 & 1.57 & 14.1 & 9.0 & -2.78 & 1.97 & 15.3 & 5.7 \\
 ISOF & -0.33 & 0.60 & 20.0 & 29.2 & -0.46 & 2.36 & 34.1 & 34.6 & -1.07 & 4.25 & 34.1 & 31.2 & -0.32 & 5.07 & 40.0 & 25.6 \\

\bottomrule
\end{tabular}
}
\captionof{table}{Change in Macro ($\Delta F_1$, in \%) and the percentage of rejected incorrect predictions for the OOD datasets.}
\label{tab:f1_improvement_ood}
\end{minipage}

The corresponding tables (Tables~\ref{tab:f1_improvement}–\ref{tab:f1_improvement_ood}) provide the full details, reporting mean values and standard deviations across folds.  
As before, we highlight the best mean and underline results that are not significantly different from the best ($p=0.05$).  
For example, with an average in-domain accuracy of 81.8\% across languages, a random uncertainty estimator would detect 19.2\% incorrect predictions.  
By contrast, rejecting only 1\% of the most uncertain predictions with the best methods captures 58.0\% of errors on average, peaking at 66.7\% for Arabic.  
This translates into a modest $+0.60$ improvement in F1.  
At a 10\% rejection rate, the overall gain grows to $+3.4$ F1.  

The SR method performs reasonably in-domain at very low rejection rates (1–5\%), but fails to scale to higher thresholds and degrades sharply in OOD scenarios, where its MC-Dropout equivalents prove more reliable.  
In our experiments this is especially clear for under-resourced languages (Hindi in-domain, Catalan out-of-domain), where SR becomes unstable and SMP/ENT-MC consistently emerge as the most reliable uncertainty estimation methods.  
Our results highlight an important contrast: good uncertainty discrimination, as measured by standard metrics, does not always translate into reliable performance gains when using abstention.  
In particular, methods like LOF and ISOF often score highly in aggregate evaluation, yet their selective prediction behaviour is unstable across languages and domains.  
Conversely, simpler baselines such as SR and ENT, while less impressive in raw metrics, prove more dependable when applied to real selective prediction.  
This discrepancy underscores the need to go beyond traditional UE metrics when evaluating practical usefulness.